\documentclass[grid,balance,upint,subscriptcorrection,varvw,mathalfa=cal=euler,spanish,french,greek,russian,vietnamese,colorlinks]{asmeconf}

\hypersetup{%
	pdfauthor={Maryam Ghasemzadeh},                       		      
	pdftitle={ASME Conference Paper LaTeX Template},                  
	pdfkeywords={ASME conference paper, LaTeX template, BibTeX style},
	pdfsubject = {Describes the asmeconf LaTeX template with author grid},
	pdflicenseurl={https://ctan.org/pkg/asmeconf},
}

\usepackage{tikz}
\usetikzlibrary{shapes.geometric, arrows,fit,backgrounds,positioning}
\usepackage{amsmath}
\usepackage{dsfont}

\begin{document}

\ConfName{Proceedings of the \linebreak International Design Engineering Technical Conferences and Computers and Information in Engineering Conference}
\ConfAcronym{IDETC-CIE 2025}
\ConfDate{Agust 17–20, 2025}
\ConfCity{California, USA}
\PaperNo{IDETC2025-168865}


\title{NOSTRA: A Noise-Resilient and Sparse Data Framework for Trust Region based Multi-Objective Bayesian Optimization} 
%
%
%


\SetAuthors{%
	Maryam Ghasemzadeh\affil{1}, 
	Anton van Beek\affil{1}\CorrespondingAuthor{anton.vanbeek@ucd.ie}
}

\SetAffiliation{1}{School of Mechanical and Materials Engineering\\ University College Dublin\\ Dublin, Ireland}


\maketitle





\begin{abstract}

Multi-objective Bayesian optimization (MOBO) struggles with sparse (non-space-filling), scarce (limited observations) datasets affected by experimental uncertainty, where identical inputs can yield varying outputs. These challenges are common in physical and simulation experiments (e.g., randomized medical trials and, molecular dynamics simulations) and are therefore incompatible with conventional MOBO methods. As a result, experimental resources are inefficiently allocated, leading to suboptimal designs. To address this challenge, we introduce NOSTRA (Noisy and Sparse Data Trust Region-based Optimization Algorithm), a novel sampling framework that integrates prior knowledge of experimental uncertainty to construct more accurate surrogate models while employing trust regions to focus sampling on promising areas of the design space. By strategically leveraging prior information and refining search regions, NOSTRA accelerates convergence to the Pareto frontier, enhances data efficiency, and improves solution quality. Through two test functions with varying levels of experimental uncertainty, we demonstrate that NOSTRA outperforms existing methods in handling noisy, sparse, and scarce data. Specifically, we illustrate that, NOSTRA effectively prioritizes regions where samples enhance the accuracy of the identified Pareto frontier, offering a resource-efficient algorithm that is practical in scenarios with limited experimental budgets while ensuring efficient performance.
\end{abstract}



\section{Introduction}\label{Introduction}
Design problems often demand the simultaneous optimization of multiple conflicting objectives, making multi-objective optimization (MOO) a vital field of research. MOO aims to approximate the Pareto frontier, representing trade-offs where improving one objective requires compromising others while incorporating the decision maker's priorities assigned through weights \cite{lin2023many,ji2020simplified}. Among the various approaches to addressing MOO, Bayesian Optimization (BO) \cite{jones1998efficient} has become a prominent strategy for optimizing expensive-to-evaluate functions, leveraging knowledge from previous experiments to adaptively guide sampling toward the optimal solutions.

The extension of BO to multiple-objective Bayesian optimization (MOBO) involves the use of multi-output or multiple surrogate models (e.g., Gaussian processes (GP), and random forests) to approximate objective functions, and guide the search process by effectively managing the surrogate model's predictive uncertainty \cite{snoek2012practical}. Specifically, the optimization process involves the use of acquisition functions, which leverage predictions from the surrogate model to identify regions in the design space that balance the trade-off between exploring new designs and refining current designs with strong performance \cite{he2024efficient,evolution2024guided,zuhal2019comparative}. Despite MOBO's theoretical appeal, its practical implementation often faces challenges when faced with specific experimental conditions. 

Some challenges, such as the diversity of the Pareto frontier \cite{konakovic2020diversity} and incorporating decision-maker preferences \cite{ozaki2024multi}, have been addressed, strengthening MOBO as a generalizable tool for design. For example, Lukovic et al. \cite{konakovic2020diversity} proposed a diversity-guided batch sampling approach that ensures well-distributed solutions at the Pareto frontier while efficiently leveraging parallel computational resources. By avoiding redundant sampling, this method reduces optimization time and enhances solution quality. In contrast, Ozaki et al. \cite{ozaki2024multi} introduced a BO framework that focuses on adaptively estimating the preferences of decision-makers through interactive feedback, rather than identifying the entire Pareto frontier. However, the effectiveness of these methods relies on both the selected surrogate model and well-designed acquisition functions, which need to perform efficiently with data from practical scenarios. Specifically, Data from practical sources (e.g., physical experiments) often exhibit three key challenges: i) \textbf{scarcity}, meaning they are limited and expensive to obtain; ii) \textbf{sparsity}, meaning their density is insufficient to adequately represent the search space; and iii) \textbf{noisy}, meaning the variability in the experimental outcome, where repeating the same experiment may yield different results \cite{alzubaidi2023survey}.

A major challenge in data-driven optimization is the scarcity of high-fidelity data, which complicates the optimization process and limits the effectiveness of conventional MOBO methods. To mitigate this issue, Irshad et al. \cite{irshad2021expected} introduced a multi-fidelity approach that leverages low-fidelity and fast-to-evaluate simulations to identify high-performing designs while reducing the total experimental budget. However, the success of this method depends on accurately modeling the correlation between low- and high-fidelity data. When this correlation is weak, the resulting surrogate models can provide misleading predictions, ultimately undermining the effectiveness of the optimization. In addition, low-fidelity data sources may not be available for certain design problems. 

Beyond data scarcity, MOBO also faces challenges due to data sparsity, particularly in high-dimensional spaces where the curse of dimensionality degrades surrogate model accuracy. In such cases, the available data can be insufficient to adequately represent the search space, leading to inefficient optimization and difficulty in maintaining a diverse and accurate Pareto frontier. To address these issues, Daulton et al. \cite{daulton2022multi} proposed dimensionality reduction techniques that enhance surrogate model performance and improve optimization efficiency. While effective, their method assumes deterministic and noise-free data, limiting its applicability in practical scenarios where experimental uncertainty (i.e., noise) is unavoidable.
Recognizing the impact of experimental uncertainty, Daulton et al. \cite{daulton2022robust} extended MOBO method to explicitly guide the optimization process toward solutions that are both Pareto-optimal and robust to experimental uncertainty. Additionally, their work on parallelizing the Expected Hypervolume Improvement (EHVI) acquisition function \cite{daulton2021parallel} enables parallel evaluations, improving computational efficiency. However, despite these advancements, existing methods fail to systematically address the combined challenges of data scarcity, sparsity, and noise. This gap highlights the need for an approach that can effectively navigate these limitations, ensuring reliable optimization outcomes in practical applications where observations are both limited and uncertain.

\begin{figure}
    \centering
    \includegraphics[width=0.9\linewidth]{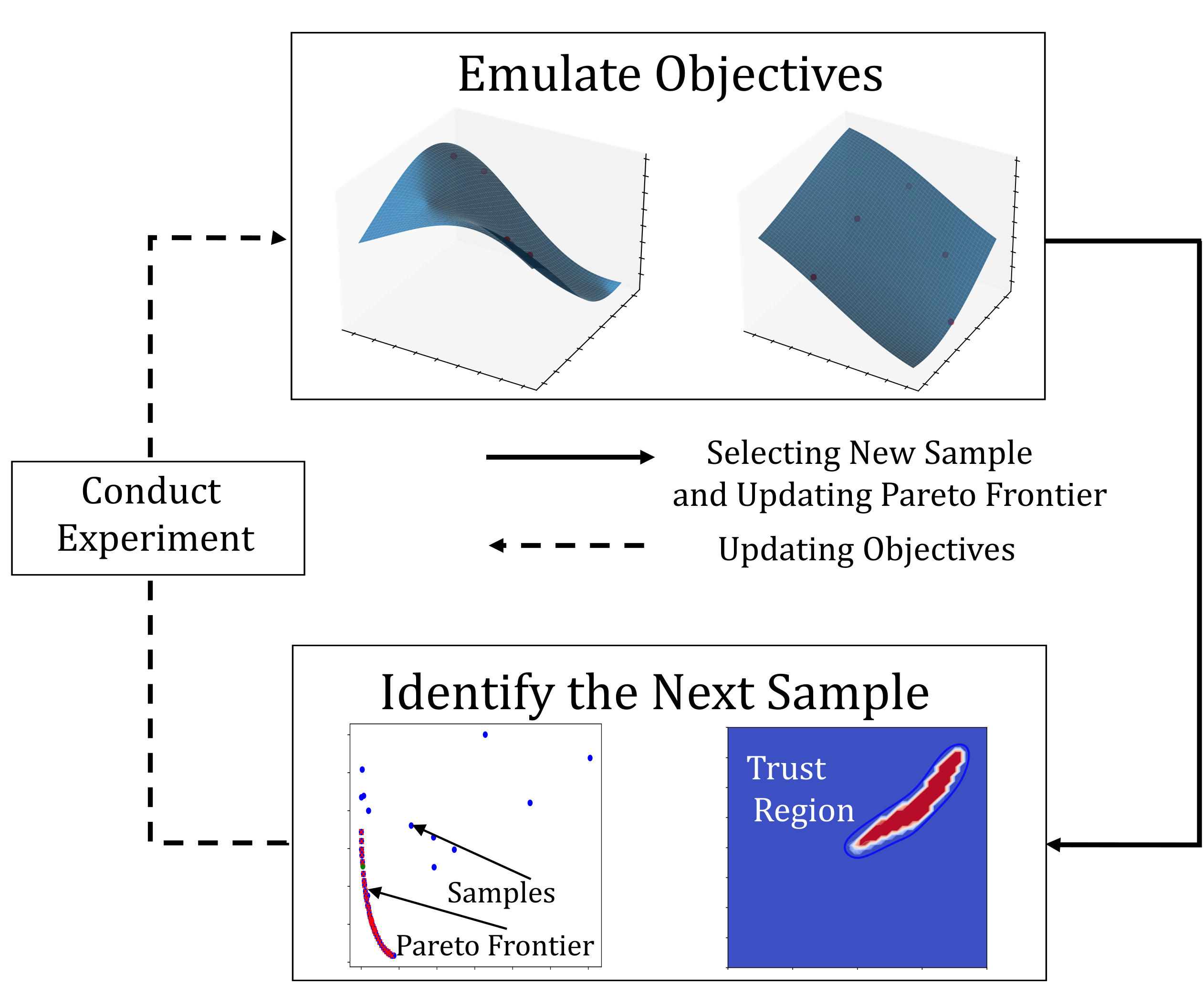}
    \caption{Trust Region Based Multi Objective Bayesian Optimization}
    \label{fig:Introductionary}
\end{figure}

In this paper, we introduce a new MOBO framework compatible with scenarios where training data is sparse, scarce, and corrupted by experimental uncertainty. The framework involves an iterative loop between constructing accurate surrogate models and identifying promising designs as shown by the iterative design loop in Figure \ref{fig:Introductionary}. Our approach begins by enhancing the surrogate model to ensure reliable performance with scarce and sparse datasets, enabling accurate predictions even under experimental uncertainty (as shown in the top panel of Figure \ref{fig:Introductionary}). Building on this, we propose an adaptive sampling strategy that focuses sample selection within designated trust regions, areas with a higher probability of containing solutions belonging to the Pareto frontier (as shown by the red region in the bottom panel of Figure \ref{fig:Introductionary}). By concentrating resources on these regions of physical or computational experiments, our method efficiently identifies a high-quality Pareto frontier using significantly fewer samples than available methods. The use of trust regions further enhances the efficiency of the algorithm by reducing resource allocation to less relevant areas, facilitating smarter exploration of the design space. While the acquisition function naturally balances exploration and exploitation, our proposed framework, NOSTRA refines this process by guiding sampling efforts toward trust regions, improving the overall efficiency and quality of the optimization process.

In contrast to prior work, NOSTRA proposes a novel trust-region-based MOBO framework that directly addresses the combined challenges of data scarcity, sparsity, and noise by operating in the output space to discretize the input space. While robust MOBO methods typically rely on explicit modeling of noise or uncertainty \cite{daulton2022robust}, NOSTRA leverages relative Pareto frontier probabilities to define adaptive trust regions. This strategy enables efficient exploration even under severely limited and noisy observational data. Moreover, unlike recent approaches that partition the design space through learned input-space discretization \cite{irshad2021expected,tao2021multi,chen2017multimodel}, NOSTRA performs clustering in the design space based on estimated Pareto membership probabilities. This allows for the dynamic and data-driven identification of high-potential regions without relying on predefined spatial partitioning. As a result, NOSTRA offers a data-efficient mechanism for guiding the optimization process, particularly in practical settings characterized by sparse and uncertain data.

The structure for the remainder of this paper is as follows: In Section~\ref{Sec:Section 2}, we introduce the theoretical foundations of GPs, which serve as the surrogate model in our framework, and provide background on MOBO. In Section~\ref{sec:Nostra}, we present our framework, NOSTRA, detailing its mathematical foundation and how it improves results by enhancing the accuracy of GP and incorporating a trust region approach. In Section~\ref{sct:numeical}, we provide a detailed implementation of the proposed method across a range of benchmark problems, accompanied by a comprehensive analysis of its efficiency and a comparison with traditional approaches. Finally, in Section~\ref{conclusion}, we summarize the key findings and present our concluding remarks.

\section{Background}\label{Sec:Section 2}

GP models are fundamental tools for modeling random processes based on the available data points (observed samples) and their correlations through a joint multivariate normal distribution. In the context of MOBO, GPs serve as surrogate models where their posterior predictive distributions facilitate balancing between exploration and exploitation when deciding what experiments to conduct next. In this section we introduce the theoretical foundations of GPs and their role in MOBO, highlighting their effectiveness in optimizing multiple conflicting objectives, particularly in data-scarce and noisy environments \cite{tao2021multi,bostanabad2018leveraging,vanbeek2021scalable}.

\subsection{Gaussian Process Models: An Overview}\label{Sec:subsection 2.1.1}
The standard formulation of GPs to approximate a single objective is defined as
\begin{equation}
y(\boldsymbol{x})=f(\boldsymbol{x})+Z(\boldsymbol{x}),
\label{eq:Gp}
\end{equation}
where $y(\boldsymbol{x})$ represents a noise corrupted output observed for a $D$-dimensional input  $\boldsymbol{x} = \{x_1, x_2, \ldots, x_D\}^T$, $f(\boldsymbol{x})$ is a mean function that capture the general trend of a process, and $Z(\boldsymbol{x})$ is a stationary stochastic process with zero mean. With the assumption that the prior mean of the process is zero ($f(\boldsymbol{x})=0$), the ordinary GP formulation is simplified to its stochastic process, $Z(\boldsymbol{x})$. The covariance function $cov(\cdot)$, also referred to as the kernel function, is a fundamental component of GPs and is defined as a measure of similarity between random variables $\boldsymbol{x}$ and $\boldsymbol{x}^\prime$, given by
\begin{equation}
cov(\boldsymbol{x},\boldsymbol{x^\prime})=\sigma^2R(\boldsymbol{x},\boldsymbol{x}^\prime),
\label{eq:kernel}
\end{equation}
where $R(\cdot)$ is the correlation function, for which a common choice is the squared exponential function defined in Equation~\eqref{eq:correlation}, and $\sigma^2$ represents the prior variance.
\begin{equation}
R(\boldsymbol{x}, \boldsymbol{x}') = \exp\left\{-\sum_{i=1}^D 10^{\boldsymbol{\omega}_i} \left(x_i - x_i'\right)^2\right\},
\label{eq:correlation}
\end{equation}


The vector of parameters \(\boldsymbol{\omega}\) governs the smoothness of the correlation function and plays a critical role in defining the behavior of the GP. Often referred to as the roughness parameter, \(\boldsymbol{\omega}\) determines how effectively the GP model captures variations caused by specific input variables. Depending on the problem and the required level of precision, either isotropic or anisotropic kernels can be utilized. The key distinction lies in the treatment of \(\boldsymbol{\omega}\): isotropic kernels use the same \(\omega\) across all input dimensions, while anisotropic kernels assign a unique value to each dimension \cite{ojha2024machine}. In this study, we use isotropic kernels to provide the surrogate models with more flexibility to accurately model the provided training data. This choice minimizes the risk of over-fitting, particularly in low-dimensional cases with sparse data \cite{kanagawa2018gaussian}.

Another important concept is experimental uncertainty, which is assumed to be homoscedastic, meaning its variance remains constant across the design space. Heteroscedastic experimental uncertainty, on the other hand, would limit its applicability to sparse data problems as they require larger training data sets (\cite{vanbeek2021scalable,plumlee2014building}). Under this assumption, the observed response surface \(\boldsymbol{y}\) can be modeled as

\begin{equation}
y_i = f(\textbf{x}_i) + \epsilon_i,
\label{eq:noisy_response}
\end{equation}

where $\epsilon_i \sim \mathcal{N}(0, \delta^2),\quad i=1,\ldots,n$ represents independent Gaussian noise with constant variance $\delta^2$ that as observed from $n$ training samples. To represent this noise, it is directly incorporated into the correlation function, modifying it as follows
\begin{equation}
\boldsymbol{R}_\delta = \boldsymbol{R} + \delta^2 \boldsymbol{I},
\label{eq:noisy_correlation}
\end{equation}

where $\boldsymbol{R}$ is the original correlation matrix between $\boldsymbol{x}$ and $\boldsymbol{x}^\prime$ as it is defined in Equation \ref{eq:correlation}, and $\boldsymbol{I}$ denotes an identity matrix. The hyperparameters of the covariance matrix, $\boldsymbol{\omega}$, $\sigma^2$, and $\delta$, must be carefully estimated during GP training to ensure accurate predictions. A common approach to determine these parameters is maximum likelihood estimation \cite{fisher1922mathematical}, which optimizes the likelihood function to fit the observed data. Alternatively, designers can use full Bayesian inference\cite{box2011bayesian}, which incorporates prior knowledge to estimate parameters, or cross-validation \cite{stone1974cross}, which selects parameters based on predictive performance.
The observed data $\mathbf{y}=\{y_1, y_2, \ldots, y_n\}^T$ represent the values of the function at specific inputs $\mathbf{x}=\{\boldsymbol{x}_1, \boldsymbol{x}_2,\ldots, \boldsymbol{x}_n\}^T$, serving as the basis for maximizing the log likelihood function, as given by
\begin{equation}
L(\delta, \omega, \sigma^2) = \frac{n}{2} \log(\sigma^2) + \frac{1}{2} \log(\mathbf{\det(R_\delta)}) + \frac{1}{2\sigma^2} \mathbf{y}^\top \mathbf{R_\delta^{-1} }\mathbf{y},
\label{eq:liklihood}
\end{equation}
where \(\det(\cdot)\) denotes the determinant operator, \(\log(\cdot)\) is the natural logarithm, \(\mathbf{R_\delta}\) is the \(n \times n\) correlation matrix with elements \(R_{i,j}\) representing the correlation between inputs \(\mathbf{x}_i\) and \(\mathbf{x}_j\).

By taking the partial derivative of \(L\) with respect to \(\sigma^2\), a closed-form solution for the prior variance can be obtained	 as 
\begin{equation}
\hat{\sigma}^2 = \frac{1}{n} \mathbf{y}^\top R_\delta^{-1} \mathbf{y},
\label{eq:prior variance}
\end{equation}
eliminating \(\sigma^2\) as a hyperparameter. The optimization problem then reduces to finding the optimal values of the nugget parameter (\(\delta^2\)) and roughness parameter (\(\omega\))

\begin{equation}
\hat{\delta}, \hat{\omega} = \underset{\omega \in \Omega, 10^\delta \in \Delta}{\text{argmin}} \ L,
\label{eq:hyperparameters}
\end{equation}

where \(\Omega \in (-10, 10)^d\) is the domain for \(\omega\), and \(\Delta \in (10^{-8}, 10^1)\) is the domain for \(\delta^2\) \cite{mao2024efficient}. The mean value of the posterior prediction of the objective under study, obtained using the optimal hyperparameters for the dataset, is denoted by \(\hat{f}(\mathbf{x}, \mathbf{y})\). This represents the posterior distribution for the function \(f\), given by \(P(f \mid (\mathbf{x}, \mathbf{y}))\).

\subsection{Multi Objective Bayesian Optimization}\label{subsec:MOBO}
The BO method is typically used for optimizing functions modeled with GPs \cite{jones1998efficient} or random forests \cite{wang2020}. For problems involving multiple objectives that must be optimized simultaneously, MOBO is employed. In MOBO, the goal is to identify a set of solutions that optimize all objectives concurrently. 

When optimizing multiple functions, MOBO uses multi-objective surrogate models to identify a set of solutions that satisfy a notion of optimality with respect to all objectives. This is different from single objective BO where a single solution can be found \cite{jones1998efficient}. In MOBO, the set of optimal solutions is referred to as the Pareto frontier, which consists of non-dominated solutions. Specifically, Pareto optimality implies that improving one objective requires a compromise in at least one other objective. Mathematically, the Pareto frontier \( \mathcal{P}_n \) is defined as

\begin{equation}
\label{Paretofrontier}
\mathcal{P}_n = \{ g(\boldsymbol{x}) \mid \boldsymbol{x} \in X_n, \nexists 
  \boldsymbol{x}' \in X_n \text{ such that } g(\boldsymbol{x}') \preceq g(\boldsymbol{x}) \},    
\end{equation}

where, \( X_n = \{ \boldsymbol{x}_i \}_{i=1}^n \) represents the set of inputs, and \( \preceq \) denotes the component-wise dominance relation. The functions \( g(\boldsymbol{x}) \) represent multiple objectives, modeled as unknown functions, and is approximated using a surrogate model (e.g., using a GP as described in \ref{Sec:subsection 2.1.1}). BO iteratively refines the training data of the surrogate model(s) by adding new samples \( \boldsymbol{x}^* \) at strategic inputs. These input conditions are identified by maximizing an acquisition function, which guides the selection of \( \boldsymbol{x}^* \) to improve the approximation of the Pareto frontier. The acquisition function balances \textit{exploration} (sampling in regions with high uncertainty) with \textit{exploitation} (sampling in areas predicted to have high performance) \cite{Frazier2018}. Among available acquisition functions, a common choice is hypervolume improvement (HVI) due to its appreciable generalizability. HVI measures the improvement in the hypervolume (HV) indicator of the Pareto frontier with respect to the current set of Pareto solutions, as defined below

\subsection*{Definition 1: HVI}

The HVI of a set of new Pareto optimal solutions \( \mathcal{P}_n' \) with respect to the current set of solutions \(\mathcal{P}_n \) and a reference point \( \mathbf{r} \) is defined as

\begin{equation}
\text{HVI}(\mathcal{P}_n' | \mathcal{P}_n, \mathbf{r}) = \text{HV}(\mathcal{P}_n', \mathbf{r}) - \text{HV}(\mathcal{P}_n, \mathbf{r}),
\label{eq:hypervolume}
\end{equation}  

where \( \mathbf{r} \) is the reference point, typically chosen as the worst possible solution. It should be noted that this definition applies to minimization problems. For maximization problems, the same definition can be used by multiplying Equation~\ref{eq:hypervolume} by \(-1\)).

\subsection*{Definition 2: \(\mathbf{HV}\)}

HV of a finite approximate Pareto frontier \( \mathcal{P}_n \) given by the \( M \)-dimensional Lebesgue measure \( \lambda_M \) of the space dominated by \( \mathcal{P}_n \), with the region bounded from below with respect to some reference point \( \mathbf{r} \in \mathbb{R}^M \). The \(\mathbf{HV}\) is computed as the measure of the union of hyper-rectangles \( [\mathbf{r}, \mathbf{v}] \), where \( \mathbf{v} \in \mathcal{P}_n \) and \( [\mathbf{r}, \mathbf{v}] \) represents the hyper-rectangle with corners at \( \mathbf{r} \) and \( \mathbf{v} \).

In the context of BO, the function values at unobserved points are unknown. Therefore, we need to compute the expected hypervolume improvement (EHVI), which is the expected value of the HVI over the posterior distribution of the predicted objective function values given as

\begin{equation}
\alpha_{\text{EHVI}}(\boldsymbol{x} | \mathcal{P}_n) = \mathbb{E}[\text{HVI}(f(\boldsymbol{x}) | \mathcal{P}_n)],
\label{eq:expected Hypervolume}
\end{equation}
where \( \alpha_{\text{EHVI}}(\boldsymbol{x} | \mathcal{P}_n) \) is the acquisition function used in this work.

\begin{figure}[t]
    \centering
\includegraphics[width=\linewidth]{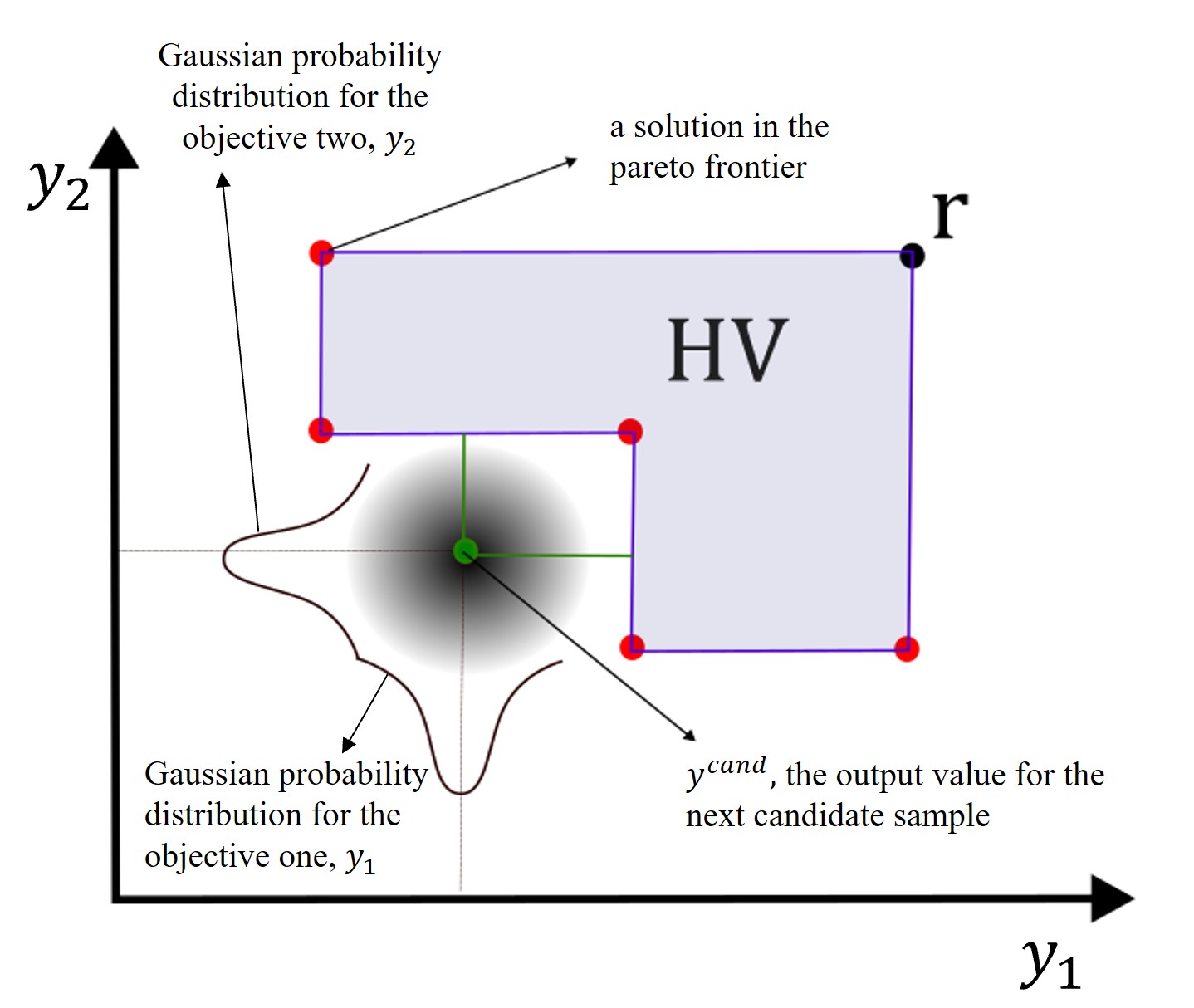}
    \caption{Illustration of calculating the Expected Hypervolume   Improvement (EHVI) for two objectives, $\mathbf{y}_1$ and $\mathbf{y}_2$}
    \label{HVI}
\end{figure}

The expectation operator can be approximated through numerical integration techniques, such as Monte Carlo (MC) sampling \cite{Emmerich2006}, as described by

\begin{equation}
\alpha_{\text{EHVI}}(\boldsymbol{x}_{\text{cand}} | \mathcal{P}_n) \approx \hat{\alpha}_{\text{EHVI}}(\boldsymbol{x}_{\text{cand}} | \mathcal{P}_n) \approx \frac{1}{N} \sum_{t=1}^{N} \text{HVI}( \hat{f}_t(\boldsymbol{x}_{\text{cand}}) | \mathcal{P}_n),
\label{eq:MontoCarlo}
\end{equation}
where \( \hat{f}_t \sim P(f | (\mathbf{x,y})) \) for \( t = 1, \dots, N \), and \( N \) is the number of MC samples. The EHVI is approximated by averaging the HVI values over multiple samples of the objective function values drawn from the GP posterior distribution. Figure \ref{HVI} illustrates the concept of EHVI for a bi-objective case involving objectives $\mathbf{y}_1$ and $\mathbf{y}_2$. This Figure demonstrates how a new candidate point contributes to expanding the dominated HV with respect to the current Pareto frontier and the reference point $r$.

\section{The NOSTRA Framework}\label{sec:Nostra}

We present advancements to the MOBO method aimed at reducing the experimental budget requirements in scenarios involving noisy, sparse, and scarce data. This section opens with a graphical introduction of NOSTRA, a comprehensive framework to highlight its two primary features: i) the use of prior knowledge on the experimental uncertainty to improve numerical stability, and ii) the use of trust regions to improve the efficiency of newly observed training samples. 

\subsection{Flowchart of the NOSTRA Framework}\label{Nostra algorithm}
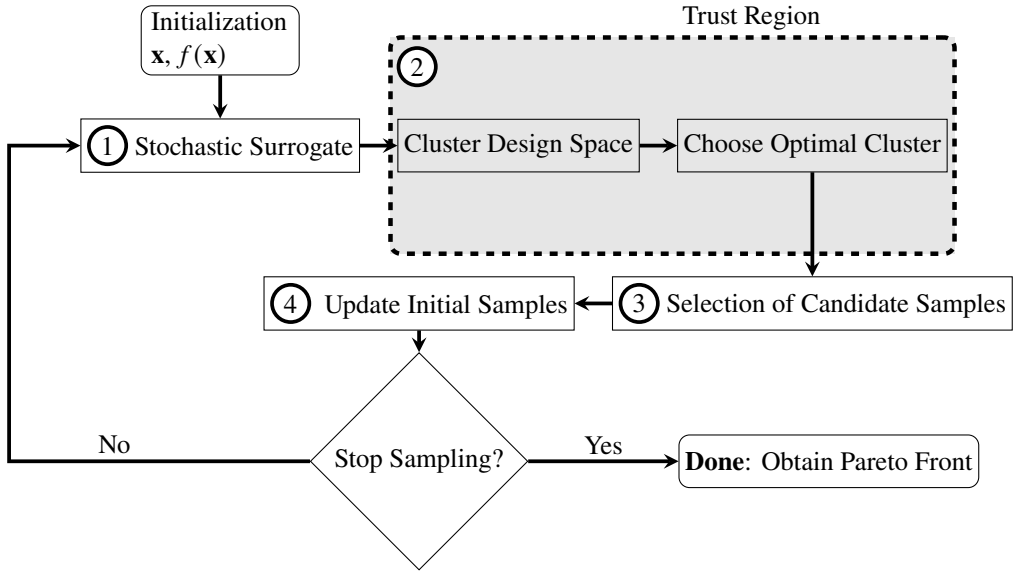
\begin{figure*}[htbp]
\centering
\begin{tikzpicture}[node distance=2cm, every node/.style={scale=0.7}]
\tikzstyle{startstop} = [rectangle, rounded corners, minimum width=3cm, minimum height=1cm,text centered, draw=black, fill=none]
\tikzstyle{Surrogate} = [rectangle, minimum width=3cm, minimum height=1cm, text centered, draw=black, fill=none]
\tikzstyle{cluster}=[rectangle,minimum width=3cm,minimum height=1cm,text centered,draw=black,fill=none]
\tikzstyle{TR}=[rectangle,minimum width=3cm,minimum height=1cm,text centered,draw=black,fill=none]
\tikzstyle{candidate}=[rectangle,minimum width=3cm, minimum height=1cm, text centered, draw=black,fill=none]
\tikzstyle{update}=[rectangle,minimum width=3cm, minimum height=1cm, text centered,draw=black, fill=none]
\tikzstyle{sampling}=[diamond,minimum width=3cm, minimum height=1cm, text centered,draw=black, fill=none]
\tikzstyle{yes}=[rectangle,rounded corners,minimum width=2cm,minimum height=1cm, text centered,draw =black,fill=none]
\tikzstyle{arrow} = [thick,->,>=stealth, line width=0.5mm]
\newcommand{\circled}[1]{
  \tikz[baseline=(char.base)]{
    \node[shape=circle, draw, line width=2pt,inner sep=2pt, fill=white,minimum size=2em] (char) {\huge #1};
  }
}
\node (start) [startstop, font=\Large, align=center] {
    \parbox{2.5cm}{
        Initialization \\ 
        $\mathbf{x}, f(\mathbf{x})$
    }
};
\node (Sur) [Surrogate, below of=start,font=\Large] {\circled{1}Stochastic Surrogate};
\node(clus)[cluster,right=0.5cm of Sur,font=\Large]{Cluster Design Space};
\node(trust)[TR,right=0.5cm of clus,font=\Large]{Choose Optimal Cluster};
\begin{pgfonlayer}{background}
    \node[fit=(clus)(trust), draw=none, fill=gray!20, inner sep=17mm] (backgroundbox) {};
    
    \node[anchor=west, xshift=0cm, yshift=1.5cm, font=\Large] at (backgroundbox.west) {\circled{2}};
\end{pgfonlayer}

\node[fit=(clus)(trust), draw, rounded corners, line width=0.6mm, dashed,inner sep=17mm, label={[font=\Large]\hspace{3cm}Trust Region}, fill=none] {};
\node(cand)[candidate,below of= trust,yshift=-1cm,font=\Large]{\circled{3}Selection of Candidate Samples};
\node(upd)[update,left=0.5cm of cand,font=\Large]{\circled{4} Update Initial Samples };
\node(samp)[sampling,below of= upd,node distance=3cm,font=\Large]{Stop Sampling?};
\node (ys) [yes,font=\Large, right=2cm of samp,node distance=5cm] {\textbf{Done}: Obtain Pareto Front};
\draw [arrow] (start) -- (Sur);
\draw[arrow] (Sur)--(clus);
\draw [arrow] (clus) -- (trust);

\draw[arrow](trust)--(cand);
\draw[arrow](cand)--(upd);
\draw [arrow] (upd) -- (samp);
\draw [arrow] (samp.west) -- ++(-4,0) node[anchor=south,xshift=2cm,font=\Large] {No} |- (Sur);
\draw[arrow](samp.east)--++(2,0) node[midway, anchor=south,font=\Large]{Yes}(ys);
\end{tikzpicture}
 \caption{Flowchart of Sequential Sampling Process.}
    \label{Fig:graph}
\end{figure*}

Figure \ref{Fig:graph} provides an overview of the NOSTRA framework, illustrating a step-by-step process for handling noisy and sparse data for efficient multi-objective optimization. We begin the algorithm by integrating the initial experimental uncertainty and limited data into a modified surrogate model, as shown by Step \scalebox{1.2}{\small\textcircled{\raisebox{-0.1em}{1}}}. This surrogate model provides an approximation of the objective functions based on the available training data. The initial training data set can be generated through an experimental design method (e.g., Latin hypercube sampling, Halton sequence, or Sobol sequence) that uniformly covers the design space with a small number of samples (e.g., $2\times D \leq n \leq 5\times D)$. 

In Step \scalebox{1.2}{\small\textcircled{\raisebox{-0.1em}{2}}}, to efficiently guide the optimization process toward high-performance designs, we adopt an adaptive trust region strategy based on clustering Pareto frontier probabilities. Specifically, for each candidate design, we estimate the probability of it being Pareto optimal. Using these probabilities, we partition the design space into distinct clusters, grouping designs with similar optimality odds. The cluster with the highest average probability of containing Pareto-optimal solutions is then selected as the trust cluster. Given the equivalence between the trust cluster and the trust region, we refer to the trust cluster as the trust region. By dedicating more samples of the experimental budget on trust regions with higher average probabilities, the search process is directed toward exploitation. This is necessary as the experimental uncertainty emphasizes exploration to reduce the posterior predictive variance to zero. However, this is unreasonable as the posterior predictive variance cannot be reduced below the intrinsic variance of the experimental uncertainty. Consequently, once the trust regions have been identified, we proceed to Step \scalebox{1.2}{\small\textcircled{\raisebox{-0.1em}{3}}} which involves selecting the next candidate sample, ensuring that the new point provides significant improvement to the approximated Pareto frontier. The process continues in Step \scalebox{1.2}{\small\textcircled{\raisebox{-0.1em}{4}}}, where the current dataset is updated with the newly observed sample. At this stage, we decide whether further sampling is necessary or if a stopping criterion has been met (e.g., the EHVI improvement is below a specific threshold, or the experimental budget has been exhausted). If additional samples are needed, the cycle repeats, gradually enhancing the accuracy of the approximated Pareto frontier while making efficient use of the available data. Through this structured approach, NOSTRA ensures data-efficient and targeted resource allocation, minimizing the experimental budget while effectively refining the set of Pareto optimal designs.

\subsection{Prior-Informed Gaussian Processes}\label{Sec:priorInfo}

\begin{figure*}[htb]
    \centering
    \begin{subfigure}{0.3\textwidth} 
        \centering
        \includegraphics[width=1\textwidth]{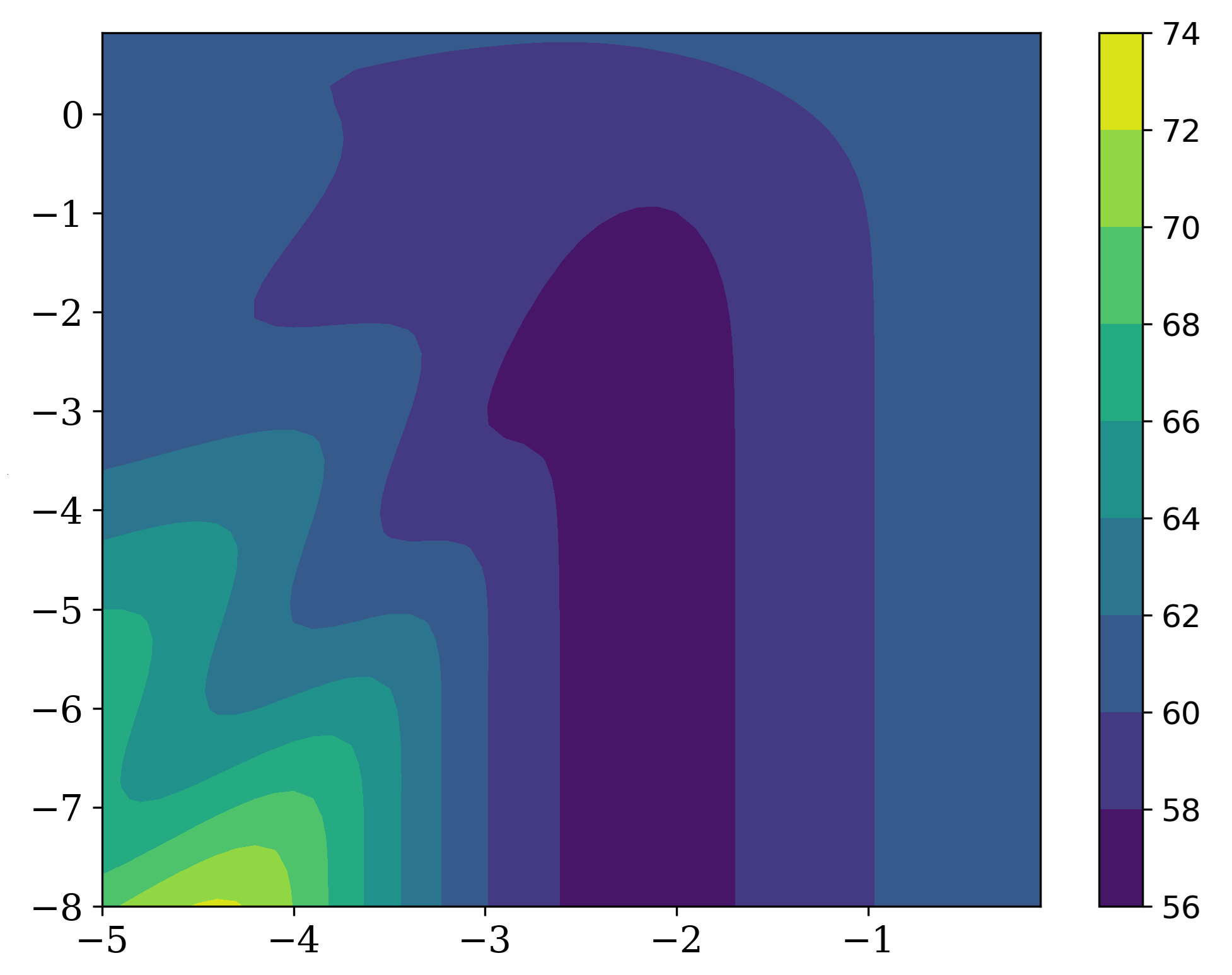} 
                \put(-90,-10){\Large{$\omega$}}
    \put(-168,70){\Large{$\delta$}}
        \caption{}
        \label{fig:logliklihood}
    \end{subfigure}
    \begin{subfigure}{0.3\textwidth}
        \centering
        \includegraphics[width=1\textwidth]{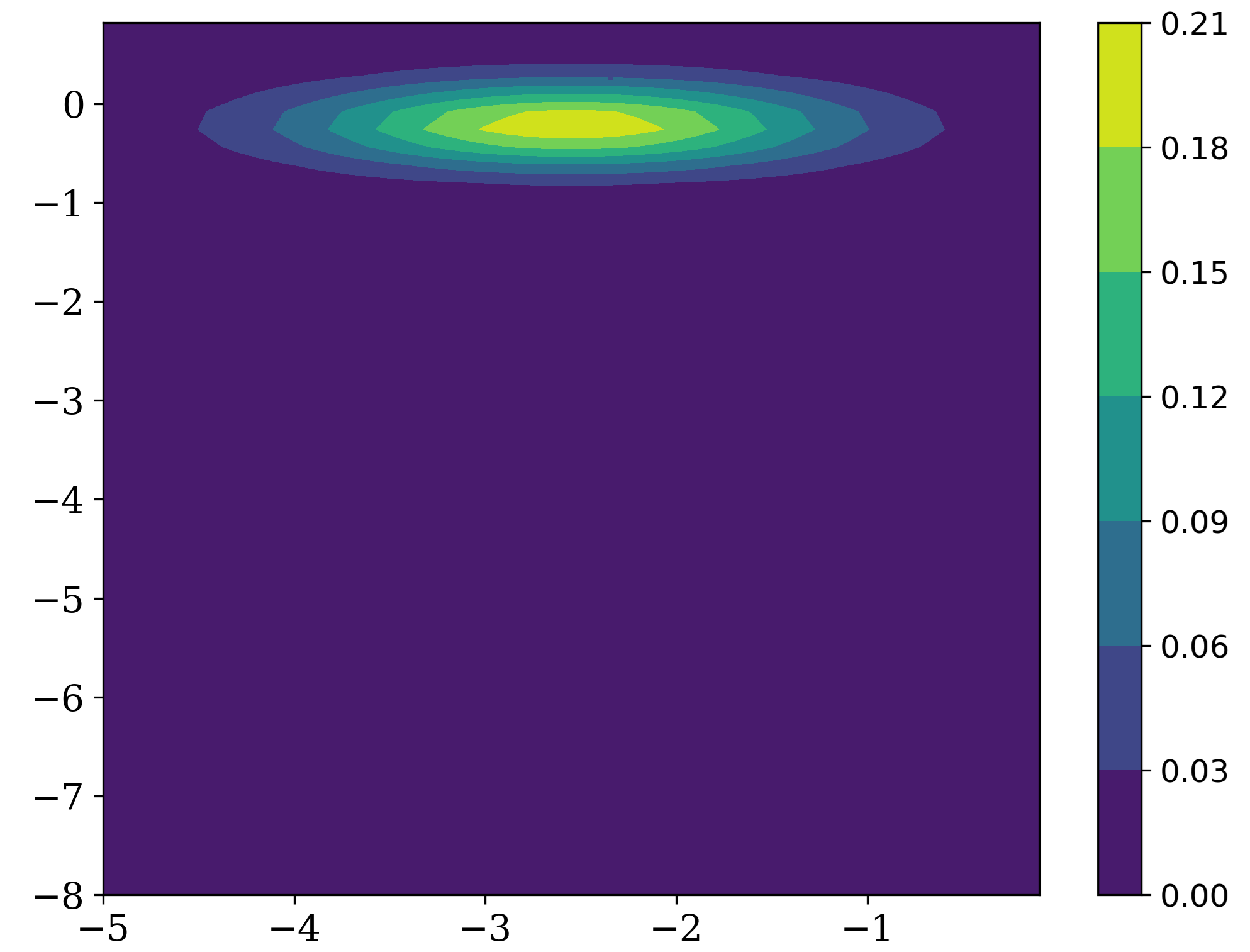} 
\put(-90,-10){\Large{$\omega$}}
        \caption{}
        \label{fig:prior}
    \end{subfigure}
    \begin{subfigure}{0.3\textwidth}
        \centering
        \includegraphics[width=1\textwidth]{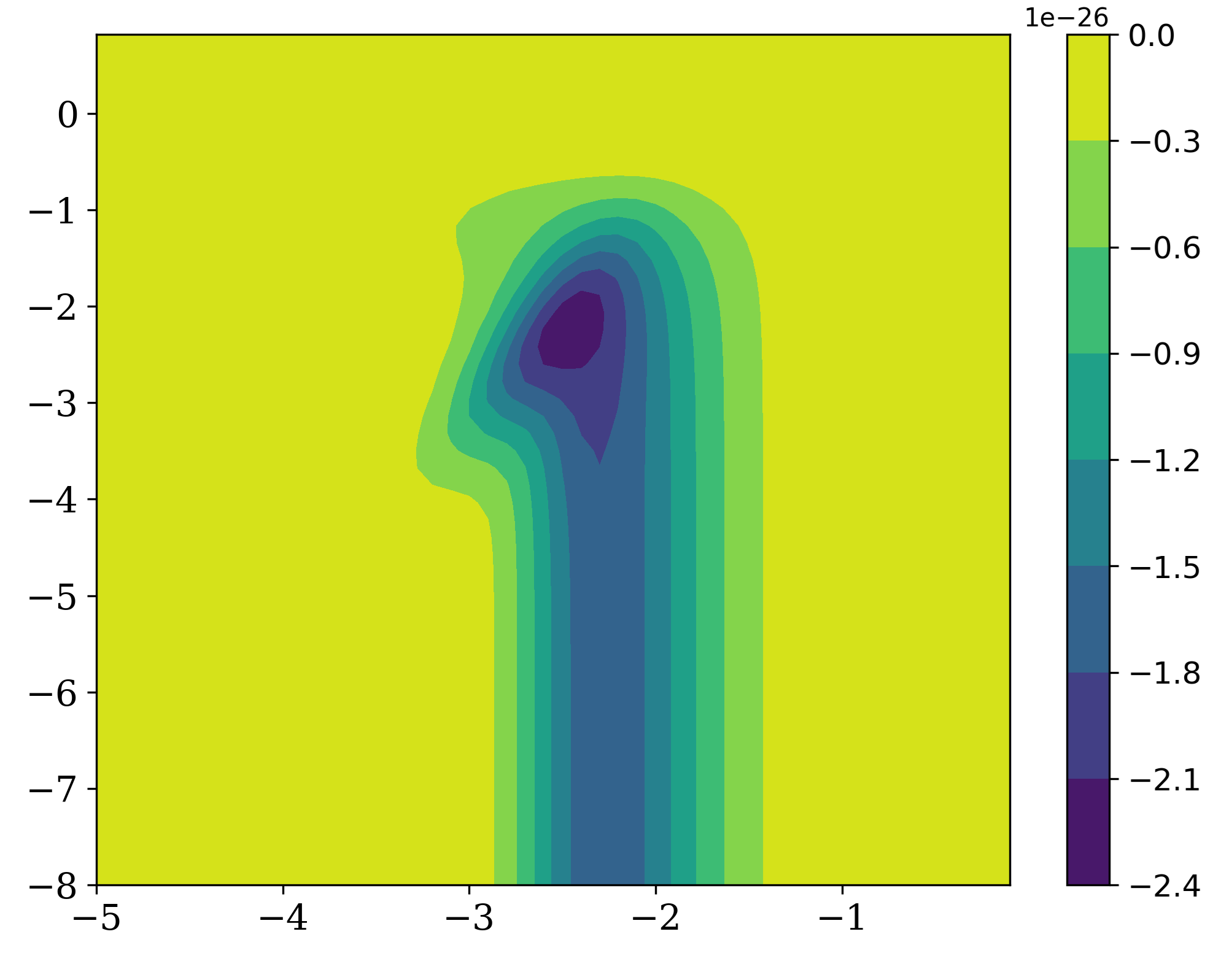}  
        \put(-90,-10){\Large{$\omega$}}
        \caption{}
        \label{fig:posterior}
    \end{subfigure}

    \caption{Bayesian Inference (a) Negative log-likelihood (b) Prior and (c) Posterior Distribution}
    \label{fig:hyperparameters}
\end{figure*}

One of the challenges that arise when using GPs with noisy, scarce, and sparse data is getting a good approximation of the hyper-parameters (i.e., $\boldsymbol{\omega}, \boldsymbol{\delta}$). This is because, under these conditions, the likelihood profile often reaches a constant and maximum value across a range of hyperparameters, resulting in a non-unique solution. This scenario has been shown through the negative log-likelihood function of the nugget parameter (\(\delta\)) and the roughness parameter (\(\omega\)) in Figure~\ref{fig:logliklihood}. The plot highlights the sensitivity of the likelihood function to these hyperparameters and reveals the absence of a unique global optimum point, making it difficult to directly determine hyperparameters by minimizing the negative log-likelihood (i.e., using maximum likelihood estimation). We address this issue by leveraging \textit{Bayes' theorem} \cite{box2011bayesian} to incorporate prior knowledge about the hyperparameters \(\omega\) and \(\delta\). Given the initially noisy and sparse data, prior information plays a crucial role in guiding the training process of our surrogate model(s). While knowledge of a function's roughness can be difficult to know a priori, we argue that engineers often have a good estimation of the signal-to-noise ratio of their experiments (e.g., such information is typically provided by suppliers of experimental equipment, or can be obtained from previous users). In addition, as more data becomes available, the likelihood progressively refines the posterior distribution even when the provided prior distribution deviates is a poor representation of reality. This can be expressed in the following equation

\begin{equation}
P(\boldsymbol{\omega}, \delta |\boldsymbol{\mathcal{D}}) \propto P(\boldsymbol{\mathcal{D}} | \boldsymbol{\omega}, \delta) P(\boldsymbol{\omega}, \delta),
\label{eq:bayes theorm}
\end{equation}
where, \(\mathbf{(x,y)=}\boldsymbol{\mathcal{D}}\) represents the training data, and \(P(\boldsymbol{\omega}, \delta |\boldsymbol{\mathcal{D}})\) is the posterior distribution of the hyperparameters, which reflects our updated beliefs after observing the data. In addition, $P(\boldsymbol{\mathcal{D}} | \boldsymbol{\omega}, \delta)$ is the likelihood of observing the data given the hyperparameters, and $ P(\boldsymbol{\omega}, \delta)$ is the prior distribution, representing our knowledge of the hyperparameters before observing any data as shown in Figure \ref{fig:prior}. When the initial data is corrupted by noise and is sparsely distributed, defined as fewer than \(10\times D\) samples \cite{rajput2023evaluation}, then combining prior knowledge with the likelihood allows us to reliably estimate the hyperparameters. For example,  in Figure~\ref{fig:posterior}, the negative posterior distribution of the hyperparameters has a unique global optimum that can be minimized using numerical solvers. 
In our implementation, the GP hyperparameters are estimated using maximum a posteriori (MAP) estimation \cite{Rasmussen2006}, which involves maximizing the log-posterior distribution rather than the marginal likelihood alone. For our subsequent test problems, we explicitly incorporate prior beliefs about the hyperparameters, assuming a normal prior distribution, as illustrated in Figure~\ref{fig:prior}. This approach enables us to guide the estimation process using prior knowledge while retaining the flexibility of data-driven inference. This is mathematically equivalent to minimizing the negative log-likelihood of the data, regularized by the log-prior, as shown in Equation ~\eqref{eq:bayes theorm} and illustrated in Figure \ref{fig:posterior}. To ensure efficient optimization, even under conditions of sparse and noisy data, we employ the multi-start L-BFGS algorithm \cite{TestFunctions}, a quasi-Newton method well suited for optimizing smooth, high-dimensional objective functions.

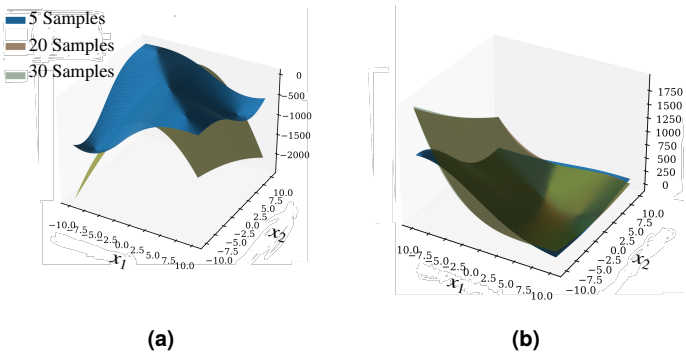
\begin{figure}[htb]
    \centering
    \begin{subfigure}{0.22\textwidth}
        \centering
        \includegraphics[width=\linewidth]{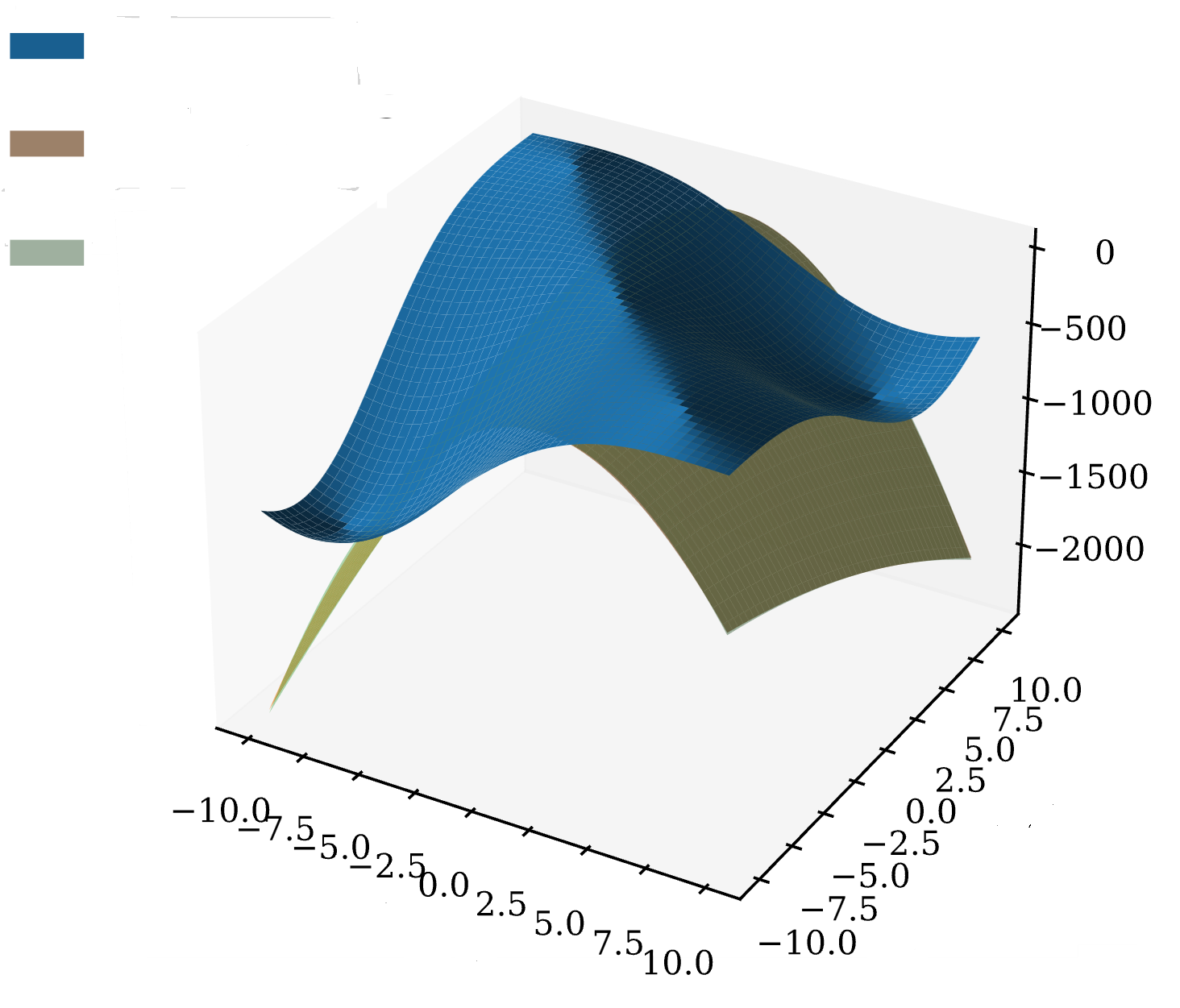}
         \begin{picture}(0,0)
        \put(-50,108){\scriptsize 5 Samples}  
        \put(-50,98){\scriptsize20 Samples} \put(-50,88){\scriptsize30 Samples} 
   \put(-20,18){\scriptsize \rotatebox{-15}{$x_1$}}
    \put(40,28){\scriptsize \rotatebox{-15}{$x_2$}}
    \end{picture}
        \caption{}
        \label{fig:Currin converge}
    \end{subfigure}
    \hfill
    \begin{subfigure}{0.22\textwidth}
        \centering
        \includegraphics[width=\linewidth]{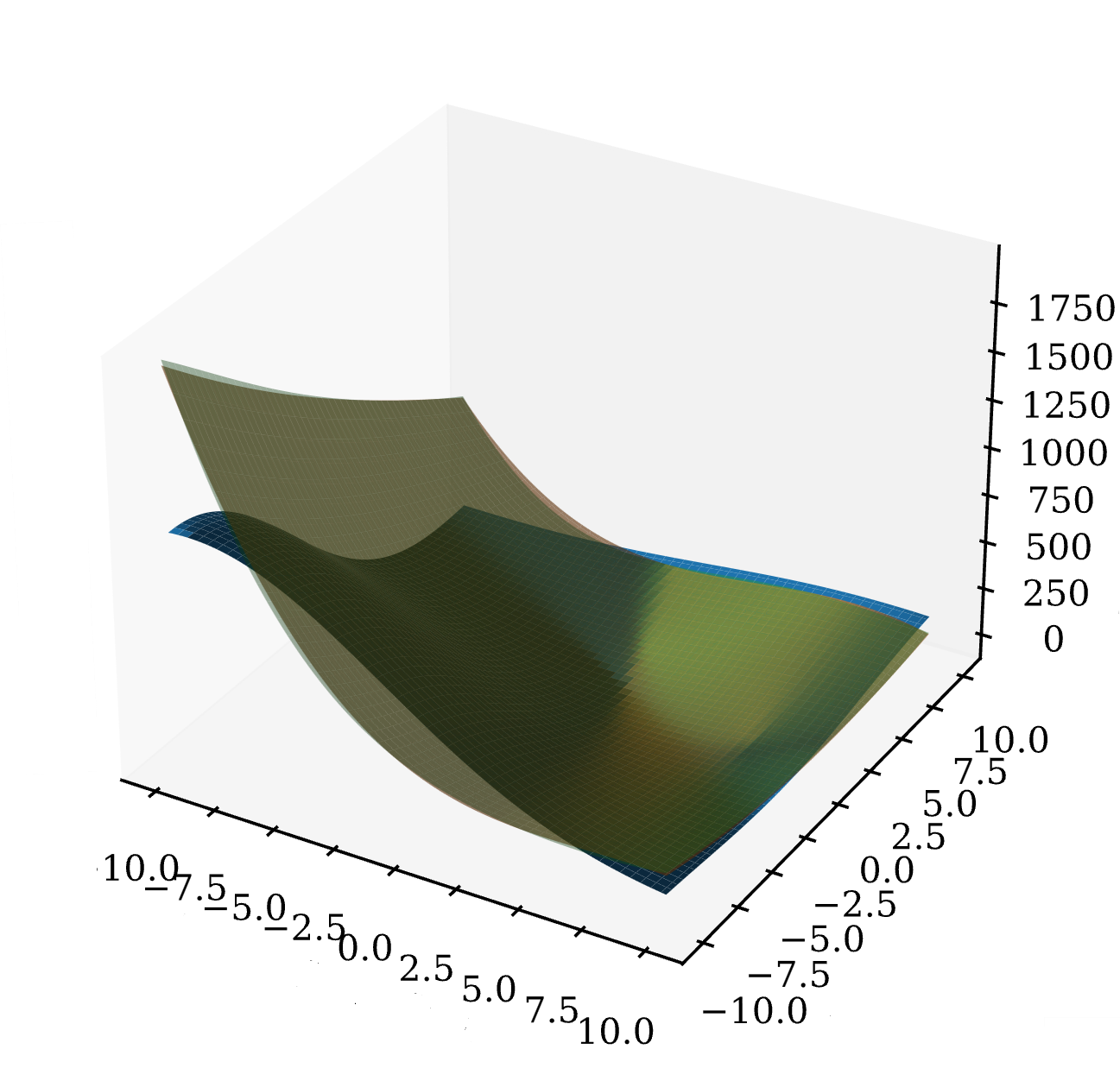}
        \put(-90,9){\scriptsize \rotatebox{-15}{$x_1$}}
        \put(-20,18){\scriptsize \rotatebox{-15}{$x_2$}}
        \caption{}
        \label{fig:Branin Converge}
    \end{subfigure}
    \caption{Efficiency of the Modified Surrogate Model:(a) Currin Function, (b) Branin Function.}
    \label{fig:convergence}
\end{figure}

We demonstrate the effectiveness of using MAP on two benchmark functions: the Currin exponential function as shown in Figure ~\ref{fig:Currin converge} and the Branin-Hoo function as shown in Figure ~\ref{fig:Branin Converge} (the mathematical formulation of these functions are provided in Section~4). These functions serve as test cases for evaluating optimization techniques due to their well-defined properties and known complexities. The dataset $\boldsymbol{\mathcal{D}}= \{\boldsymbol{x}_i, y_i\}_{i=1}^n$ where $n$ is the number of training data consists of input-output pairs along with the noise variance \(\delta\), where \(n= 5, 20, 30\) represents the number of samples used in the proposed surrogate model. The number of samples significantly impacts the quality of function reconstruction. As shown in Figure \ref{fig:convergence}, there is a visible improvement in reconstruction accuracy when the sample size increases from 5 to 20. However, further increasing the sample size to 30 shows minimal difference, as the reconstructed function with 20 samples closely resembles the approximation with 30 samples. For this analysis, the standard deviation of the normally distributed noise \(\delta^2\) is set to 5\% of the range of each function.

\subsection{Identifying and Weighing Trust Regions}\label{subsec:trustregionsampling}

In Section \ref{subsec:MOBO}, we highlighted the importance of the acquisition function in balancing exploration with exploitation for efficient resource allocation. Striking this balance is crucial to prevent convergence to suboptimal solutions, which can occur with excessive exploitation and or exploration. Overemphasizing exploitation risks clustering samples in a single region, potentially overlooking other regions that could yield better designs \cite{Rasmussen2006,daulton2021parallel}. Conversely, exploration is particularly beneficial in the early stages of the optimization process, prioritizing it too heavily can lead to slow convergence and inefficient use of resources \cite{snoek2012practical}. This challenge is especially pronounced when data acquisition is expensive and the posterior predictive distribution is inflated due to experimental uncertainty, underscoring the importance of strategic resource allocation and avoiding the exploration of regions with low potential solutions. Our approach, NOSTRA, balances exploration and exploitation by adaptively prioritizing areas of the search space based on their Pareto optimality probability when the available data is noisy and sparse. Trust regions have been used in MOBO to improve sampling efficiency for deterministic functions \cite{daulton2022multi}. Here we will explore the use of trust regions in combination with objective functions corrupted by experimental uncertainty, by partitioning the design space into sub-regions that contain designs with comparable probabilities of being a part of the Pareto frontier. 

These probabilities are approximated by MC sampling realizations of the GP posterior predictive distribution(s) and recording the frequency with which a design is a part of the Pareto frontier. Specifically, for a space-filling set of candidate samples \( \mathbf{x}_i \in \mathbf{X},\quad i=1,\ldots,M  \), we approximate the objectives \( k = 1, \ldots, K \) (where \( K \) is the number of conflicting objectives) using GP models. The posterior predictive distribution for the \( k \)-th objective at \( \mathbf{x}_i \) is given by

\begin{equation}
\label{kth posterior}
\hat{f}_k(\mathbf{x}_i) \sim \mathcal{N}(\mu_k(\mathbf{x}_i), \sigma_k^2(\mathbf{x}_i)),
\end{equation}

where \( \mu_k(\mathbf{x}_i) \) and \( \sigma_k^2(\mathbf{x}_i) \) are the predicted mean and variance for the $i-th$ candidate sample, respectively.

For the entire set of candidate samples \( X\), we draw $N$ realizations of the output \( y_{k,i}^j \) from the posterior predictive distribution of the \( k \)-th objective as

\begin{equation}
\label{sampling}
y_{k,i}^j \sim \mathcal{N}(\mu_k(\mathbf{x}_i), \sigma_k^2(\mathbf{x}_i)).
\end{equation}
We recommend setting the number of candidate samples to be relatively large to adequately cover the space of admissible designs (e.g., $M >> 100\times D)$.

By sampling \( N \) times for all \( \mathbf{x}_i \in \mathbf{X} \), we obtain \( N \) approximated Pareto frontiers, denoted as \( \mathcal{P}_k \) for \( k = 1, \ldots,K \). Each approximated Pareto frontier is constructed based on the dominance relationship among the sampled candidate points \( \mathbf{y}^j \), as defined in Equation \ref{Paretofrontier}.

Assuming a probability density function \( p(\mathbf{y}) \) for the samples generated by the GP, the expected frequency of \( \mathbf{x}_i \) being on the Pareto frontier is

\begin{equation}
\label{eq:Probability}
\mathbb{E}[c(\mathbf{x}_i)] = \int_{\mathbf{y}} \mathds{1}\{\mathbf{x}_i \in \mathcal{P}(\mathbf{y})\} \, p(\mathbf{y}) \, d\mathbf{y},
\end{equation}

Here, \( \mathds{1}\{\mathbf{x}_i \in \mathcal{P}(\mathbf{y})\} \) is an indicator function that equals $1$ if the corresponding realizations of \( \mathbf{x}_i \), \( \mathbf{y} \), lie on the Pareto frontier \( \mathcal{P}(\mathbf{y}) \), and 0 otherwise. The term \( p(\mathbf{y}) \) represents the joint probability density function of the samples \( \mathbf{y} \) as defined by the GP model, while \( d\mathbf{y} \) denotes the differential element over the sample space of \( \mathbf{y} \). The integral in the Equation \ref{eq:Probability} can be approximated using MC sampling, as

\begin{equation}
\label{eq:Monto carlo}
\mathbb{E}[c(\mathbf{x}_i)] \approx \frac{1}{N} \sum_{j=1}^N \mathds{1}\{\mathbf{x}_i \in \mathcal{P}(\mathbf{y}_j)\},
\end{equation}

where \( \mathbf{y}_j \) represents the \( j \)-th sampled set of candidate points, and \( N  \) is the total number of samples.

\begin{figure*}[htbp]
    \centering
        \begin{subfigure}{0.33\textwidth}
        \centering
        \includegraphics[width=\textwidth]{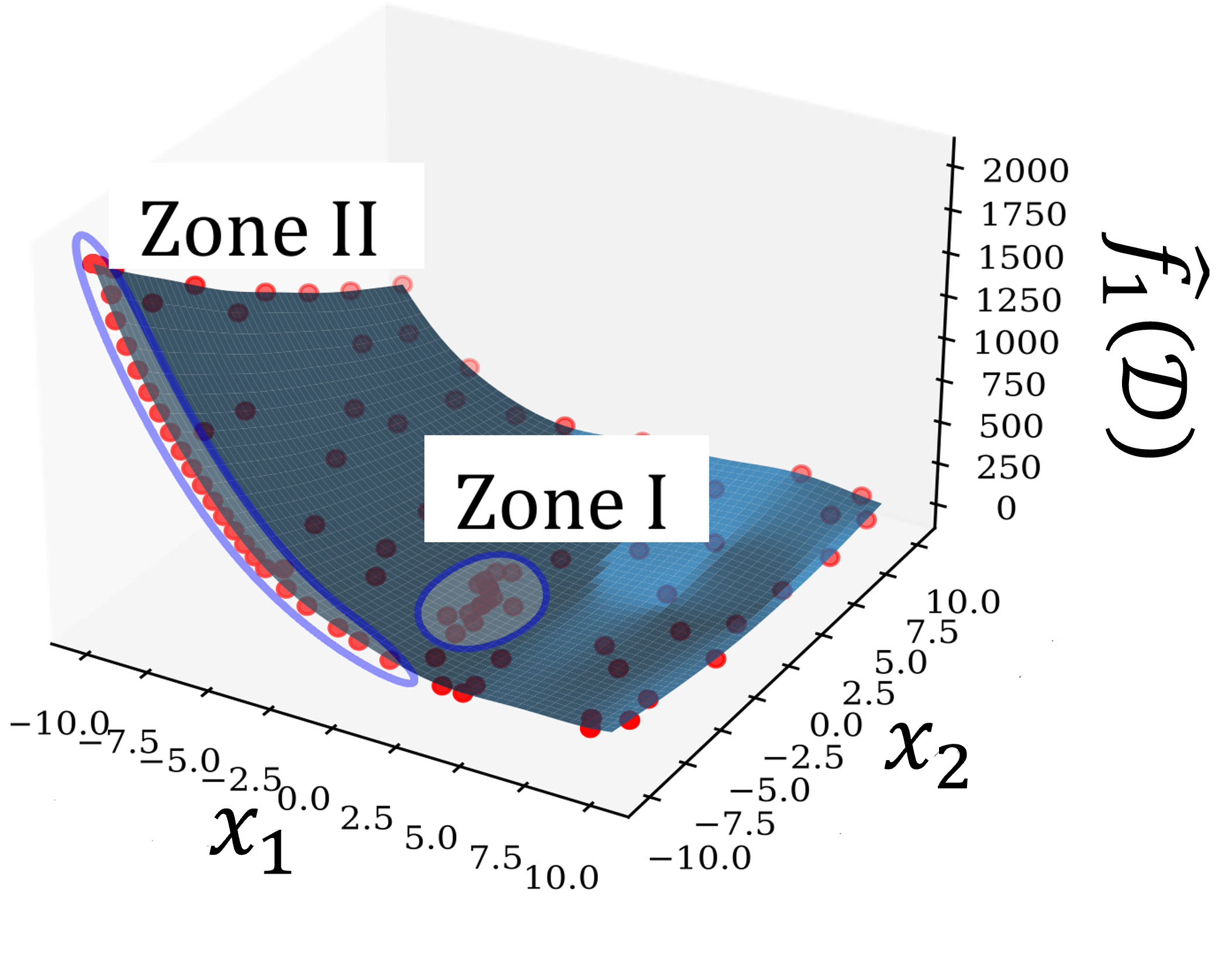}
        \caption{}
        \label{fig:braninZone}
    \end{subfigure}
    \hfill
     \begin{subfigure}{0.33\textwidth}
        \centering
        \includegraphics[width=\textwidth]{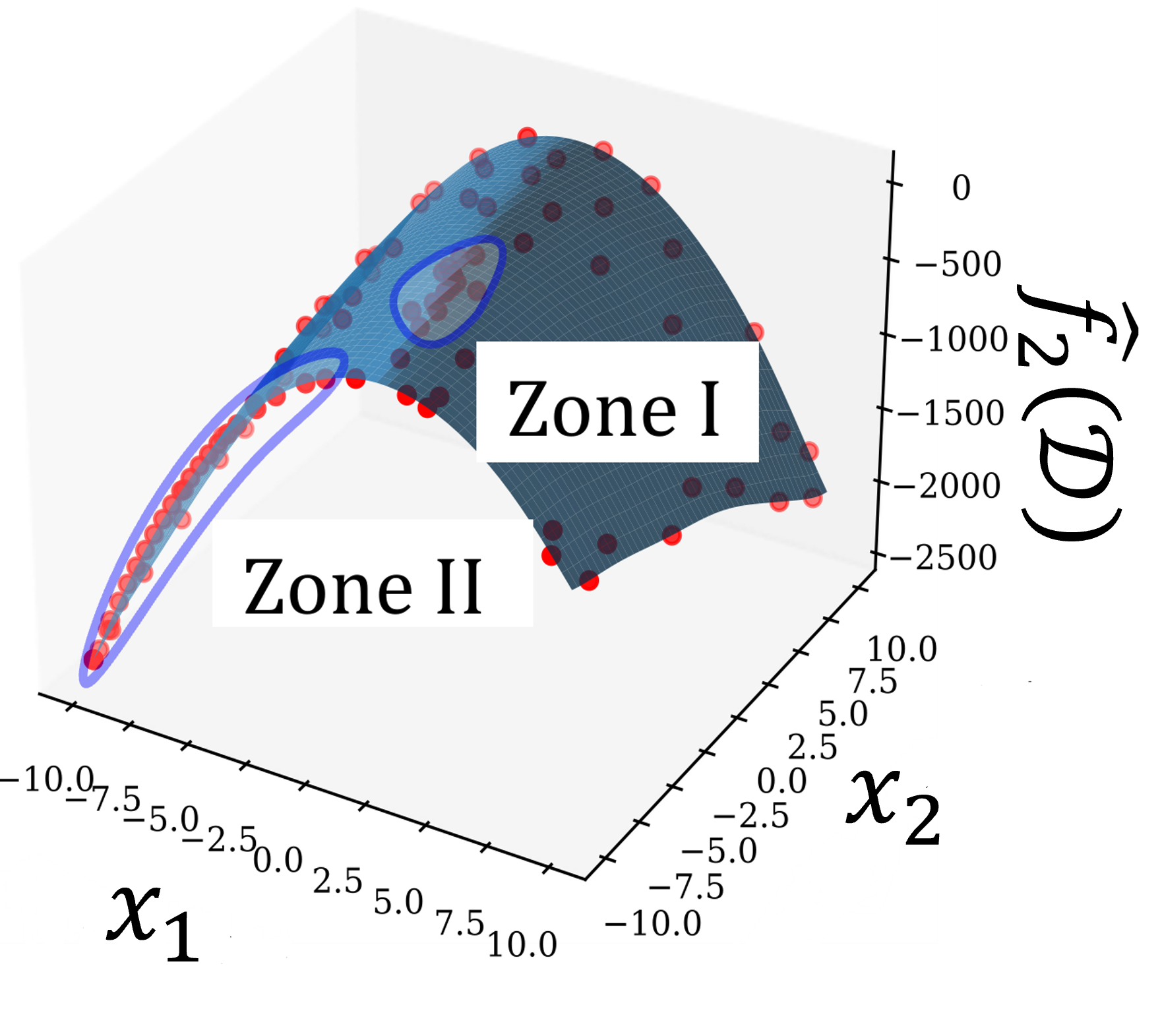}
        \caption{}
        \label{fig:CurrinZone}
    \end{subfigure}
    \hfill
      \begin{subfigure}{0.33\textwidth}
        \centering
        \includegraphics[width=\textwidth]{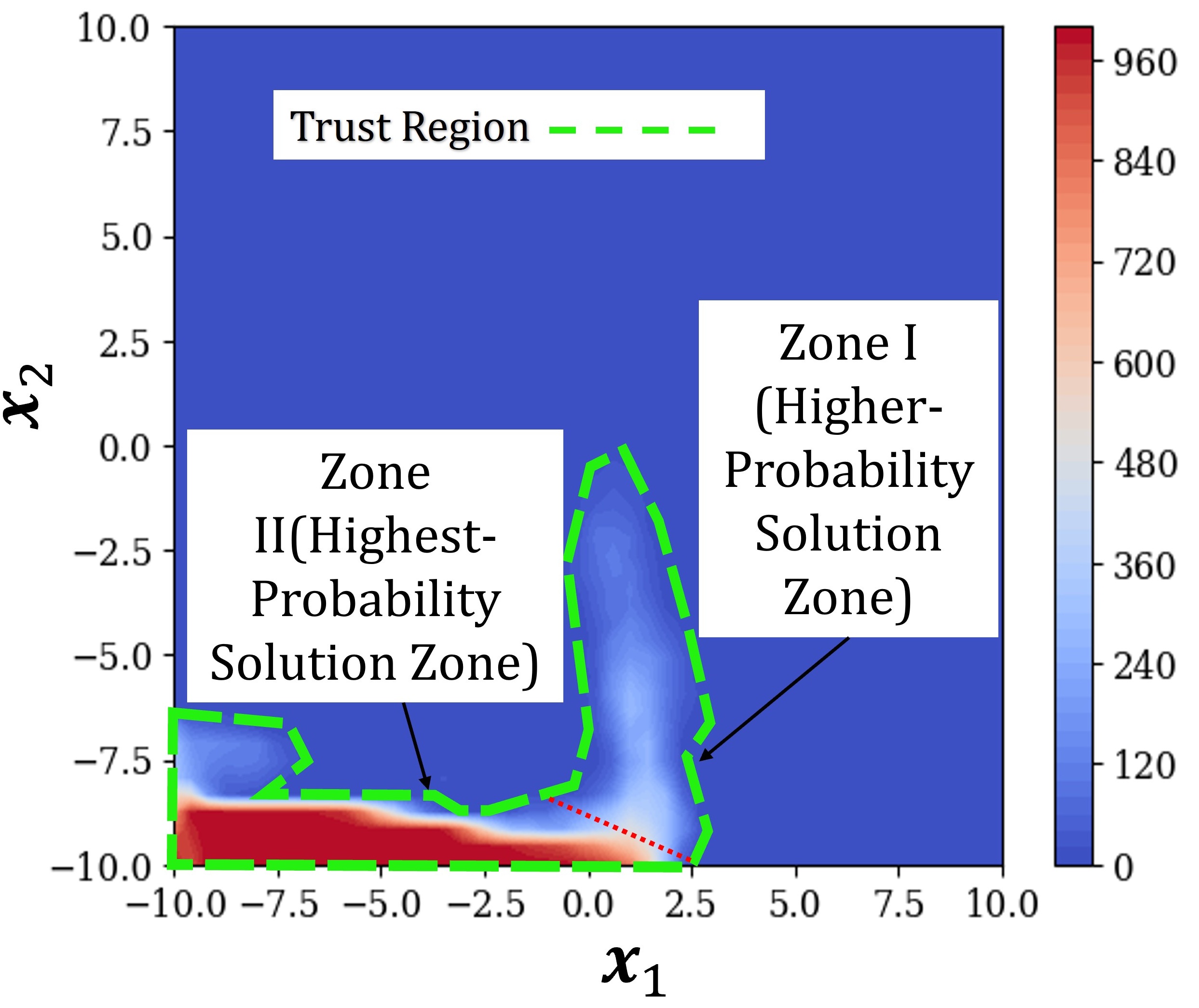} 
        \caption{}
        \label{fig:trusted}
    \end{subfigure}
    \caption{Illustration of Trust Region Sampling in NOSTRA  (a) Trust region’s zones constructed over the Branin function, highlighting areas with high sampling potential. (b) Trust region’s zones constructed over the Currin function, highlighting areas with high sampling potential.  (c) Trust region visualization showing how NOSTRA focuses sampling within promising subspaces, where probability of response guides the shape and size of the exploration zones of the trust region.}
    \label{fig:mainTrustet Region}
\end{figure*}

Continuing the test problem introduced in Subsection~\ref{Sec:priorInfo}, Figure~\ref{fig:trusted} shows how often each design point appears on the Pareto frontier for minimizing the Branin and Currin functions \cite{TestFunctions}. The region enclosed by the dashed line exhibits a noticeably higher concentration of Pareto frontier probabilities compared to the rest of the design space and is therefore identified as a trust region. Within this trust region, distinct zones emerge based on variations in Pareto probabilities, reflecting different trade-offs between the conflicting objectives. These zones provide a more detailed view of the optimality landscape and can guide more focused sampling strategies.
These distinct zones can be identified based on the trade-offs between the conflicting objectives. As shown in Figure \ref{fig:trusted}, \textbf{Zone I} falls within the trust region and has a high probability of containing Pareto-optimal solutions. However, in this zone, the Branin and Currin functions conflict significantly, with the minimum of the Branin function often coinciding with the maximum of the Currin function. Despite its relatively higher probability, the conflicting nature of the objectives is well identified through \textbf{Zone I}. Conversely, \textbf{Zone II} exhibits a less significant contrast between the two functions. Here, the interplay between the objectives is more balanced, making \textbf{Zone II} a promising region for selecting Pareto-optimal samples.

While \textbf{Zone I} has a low probability of yielding better designs, it should not be entirely neglected. Consequently, a systematic way is required to allocate weights to the different trust regions. In addition, it could be desirable to divide the design space into more trust regions. These challenges can be addressed through the use of a systematic clustering algorithm.

\subsection{Clustering Techniques for Dividing the Design Space into Trust Regions}
\begin{figure}[htb]
    \centering
    \begin{subfigure}{0.22\textwidth}
        \centering
        \includegraphics[width=\linewidth
        ,height=4cm]{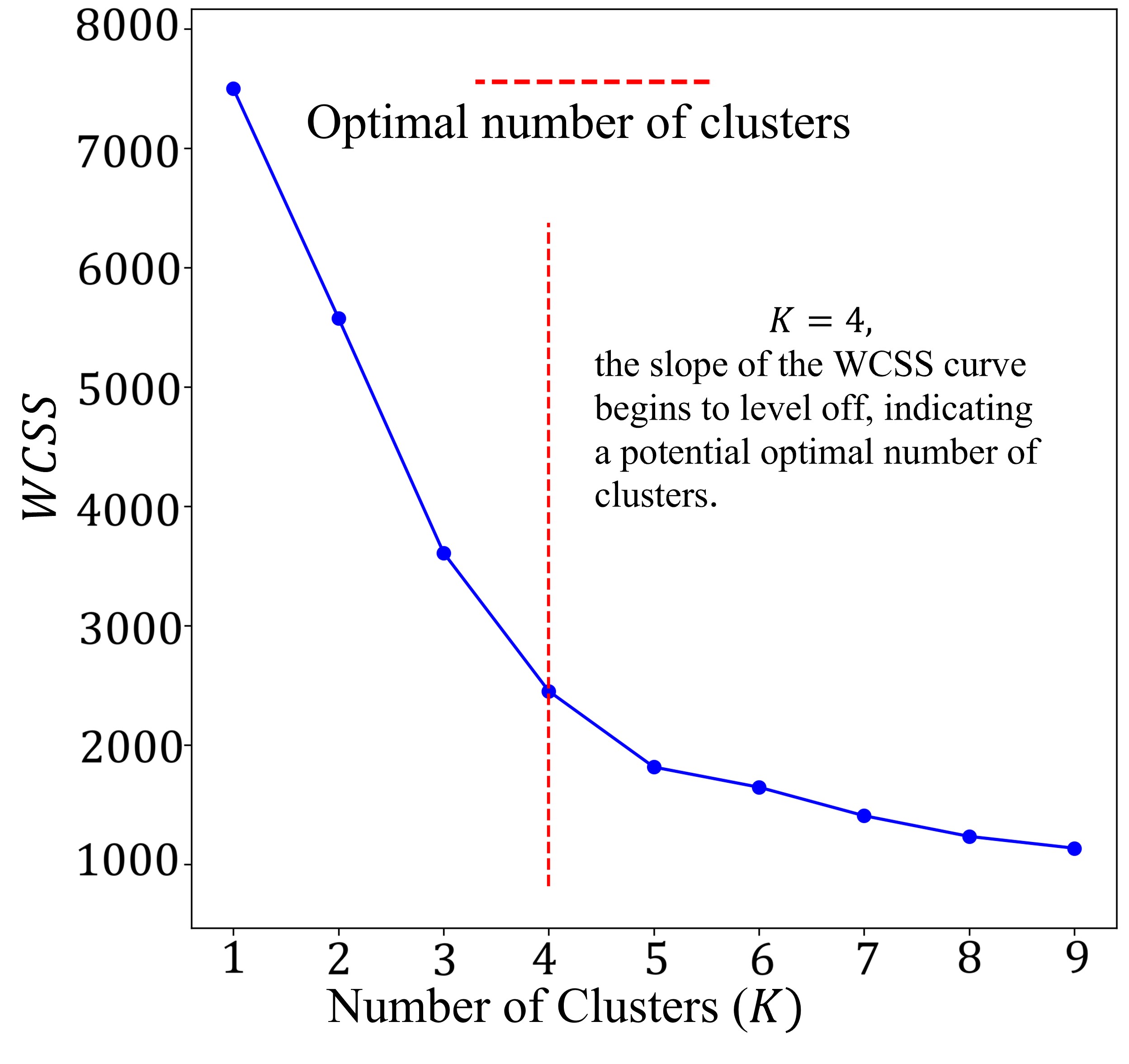}
        \caption{}
        \label{fig:elbow}
    \end{subfigure}
    \begin{subfigure}{0.22\textwidth}
        \centering
        \includegraphics[width=\linewidth,height=4cm]{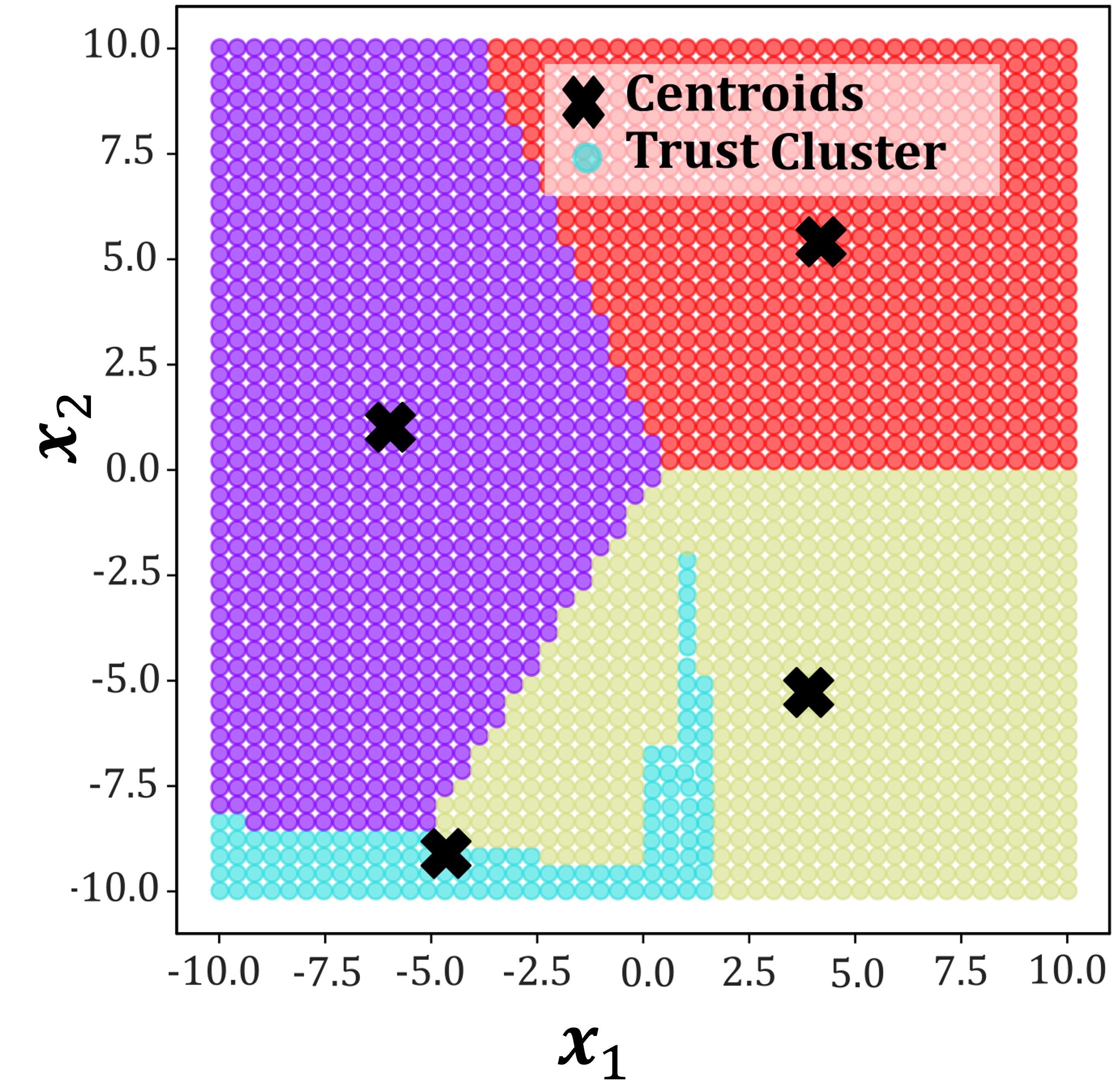}
        \caption{}
        \label{fig:cluster}
    \end{subfigure}
    \caption{Design Space Partitioning and Cluster Identification (a) Elbow plot showing the optimal number of clusters determined using the Within-Cluster Sum of Squares (WCSS) method. (b) Visualization of the design space partitioned into distinct clusters based on the identified structure.}
    \label{fig:clustering method}
\end{figure}

To partition the design space into regions with a high probability of belonging to the Pareto frontier, we employ the $K$-means clustering algorithm, a widely used unsupervised machine learning method \cite{kodinariya2013review}.
The probability of each design point belonging to the Pareto frontier is first determined using Equation~\eqref{eq:Probability}, and is then used as the input to the clustering algorithm. 
The $K$-means clustering algorithm requires the pre-specification of the number of clusters, \( K \). To determine an appropriate value for \( K \), we employ the Elbow Method \cite{hamerly2003learning,kodinariya2013review}. This approach involves balancing the \textit{within-cluster sum of squares (WCSS)} with respect to the number of clusters \( K \). Specifically, the algorithm identifies the Elbow point, which represents the optimal number of clusters by balancing the reduction of intra-cluster variance with the need to avoid excessive clustering complexity.
In Figure~\ref{fig:elbow}, we provide a visual representation of the Elbow Method for the Branin and Currin functions, corresponding to the densities shown in Figure~\ref{fig:trusted}. Initially, the algorithm shows that all samples in the design space belong to a single cluster. However, as the number of clusters increases, the WCSS decreases. However, after setting \( K = 4 \), the rate of decrease in \textit{WCSS} significantly diminishes, indicating the Elbow point and suggesting the optimal number of clusters. According to Figure \ref{fig:cluster}, one of these clusters is particularly significant for finding the Pareto frontier, as it exhibits a higher average probability of containing Pareto-optimal solutions and is therefore designated as the trust region. 

We assigned significance to each cluster by taking the product of the EHVI acquisition function with a weight, $w_{cl},\quad cl=1,\ldots,K$, determined using the equation \ref{eq:clusterWeighting}.
\begin{equation}
    w_{cl}=\frac{1}{M_{cl}}\sum_{i=1}^{M_{cl}} \mathbb{E}[c(\mathbf{x}_i)],
    \label{eq:clusterWeighting}
\end{equation}
where $M_{cl}$ represents the number of design points within each cluster, and $\mathbb{E}[c(\mathbf{x}_i)]$ denotes the expected frequency of a design point appearing on the Pareto frontier explained in Equation \ref{eq:Probability}.
Initially, the optimal number of clusters fluctuates as new samples are added, and the surrogate model progressively refines its approximation of the underlying function. This adaptive strategy differs from the use of a fixed number of clusters throughout the process. By employing the Elbow Method, the number of clusters is dynamically adjusted at each iteration, providing greater flexibility and adaptability during the optimization process.

To enhance efficiency during optimization, the number of clusters in the $K$-means algorithm is dynamically adjusted at each iteration using the Elbow Method. While the trust region delineates a broad area of promising solutions, the probability of Pareto optimality can vary significantly among design points within this region. To better capture this variation, we refine the initial coarse clustering by increasing the number of clusters $K$. The Elbow Method helps guide the selection of $K$ by identifying a balance between model simplicity and accuracy—specifically, the point where further increases in $K$ yield diminishing returns in $WCSS$ reduction (Figure \ref{fig:elbow}). For example, the method may suggest $K=4$, revealing clusters with notably higher average Pareto probabilities.
Despite identifying high-probability clusters, such as the trust region illustrated in Figure \ref{fig:cluster}, considerable local variability can still exist. This observation motivates further refinement by incrementally increasing, allowing us to zoom in on subregions (zones) with the highest likelihood of containing Pareto-optimal solutions. The Elbow Method thus provides a data-driven baseline to prevent both overfitting (too many clusters) and underfitting (clusters that are too broad). Additionally, designers may choose to manually adjust $K$ to emphasize either exploitation (larger $K$ for finer resolution) or exploration (smaller $K$ for broader coverage). We systematically analyze this trade-off in Section \ref{sct:numeical}.

\section{ Numerical Experiments}\label{sct:numeical}
To assess the performance of NOSTRA, we tested it against the widely adopted MOBO method, using EHVI as the acquisition function across two distinct test problems. We considered three NOSTRA configurations: i) a fix number of 10 trust regions, ii), a fixed number of four trust regions,  iii) a fixed number of one trust region, and iv) a dynamic number of trust regions using the Elbow method. All methods were provided with the same set of $n=2\times D$ initial input conditions. Finally, the performance of each approach is quantified through the EHVI as a new sample is added to the observed data during the optimization process.
\subsection{Test Problems}
In this study we used the following two test problems. 

\textbf{Problem 1: Bohachevsky-Sphere Function}

\[
\min_{\mathbf{x} \in \mathbb{R}^2} 
\begin{cases} 
f_1(\mathbf{x}) = (x_1 - 8)^2 + (x_2 - 8)^2 +\epsilon_i \\
f_2(\mathbf{x}) = x_1^2 + 2x_2^2 - 0.3\cos(3\pi x_1) - 0.4\cos(4\pi x_1) + 0.7+\epsilon_i
\end{cases}
\]
Both functions are subject to noise of 5\% of their respective ranges:
\[
  \epsilon_i \sim \mathcal{N}(0, (0.05 \Delta f_i)^2), \quad \forall i \in \{1, 2\},
\]
where \( \Delta f_i = \max(f_i) - \min(f_i) \) represents the range of the \( i \)-th objective function.

\textbf{Problem 2: Branin-Currin Function} 
\[
\min_{x \in \mathbb{R}^2} 
\begin{cases} 
f_1(x) = a \left( x_1 - b x_0^2 + c x_0 - r \right)^2 + s (1 - t) \cos(x_0) + s+\epsilon_i \\
f_2(x) = \left( 1 - \exp\left( -\frac{1}{2x_1} \right) \right) 
\frac{\left(2300x_0^3 + 1900x_0^2 + 2092x_0 + 60\right)}
{\left(100x_0^3 + 500x_0^2 + 4x_0 + 20\right)}+\epsilon_i 
\end{cases}
\]

Here, the constants for the Branin function are:
\[
a = 1, \, b = \frac{5.1}{4\pi^2}, \, c = \frac{5}{\pi}, \, r = 6, \, s = 10, \, t = \frac{1}{8\pi}.
\]

Both functions are subject to additive noise, where the standard deviation of the noise is initially set to be 5\% of the respective function's range.

\begin{figure*}[htb]
    \centering
    \begin{subfigure}{0.45\textwidth}
        \centering
        \includegraphics[width=\linewidth,height=7.4cm]{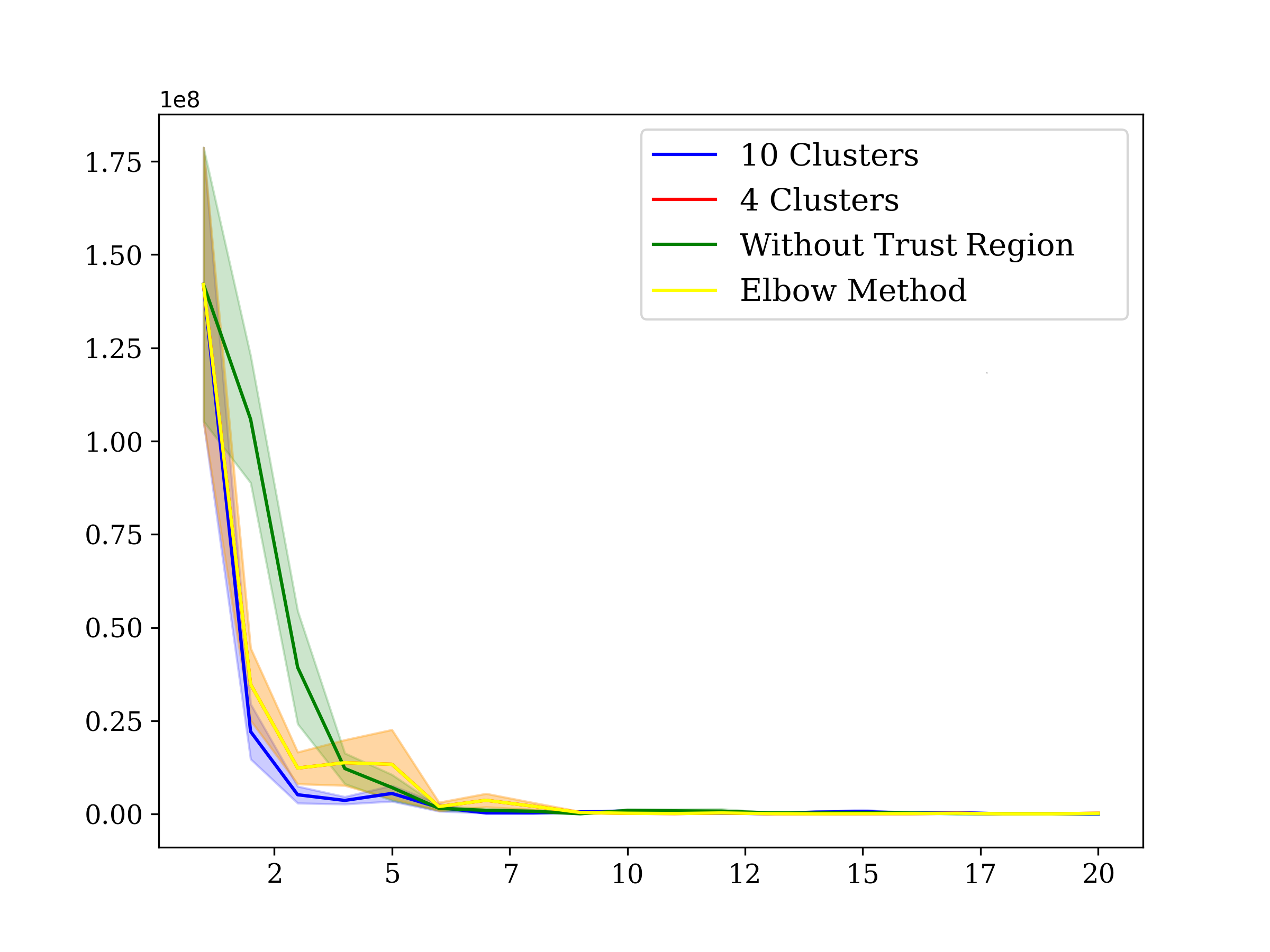}
         \put(-160,0){\normalsize\rotatebox{0}{ Design Evaluations}}
        \put(-240,80){\normalsize\rotatebox{90}{EHVI}}
        \caption{}
        \label{fig:bohachovke}
    \end{subfigure}
        \begin{subfigure}{0.45\textwidth}
      \centering
    \includegraphics[width=\linewidth,height=6.8cm]{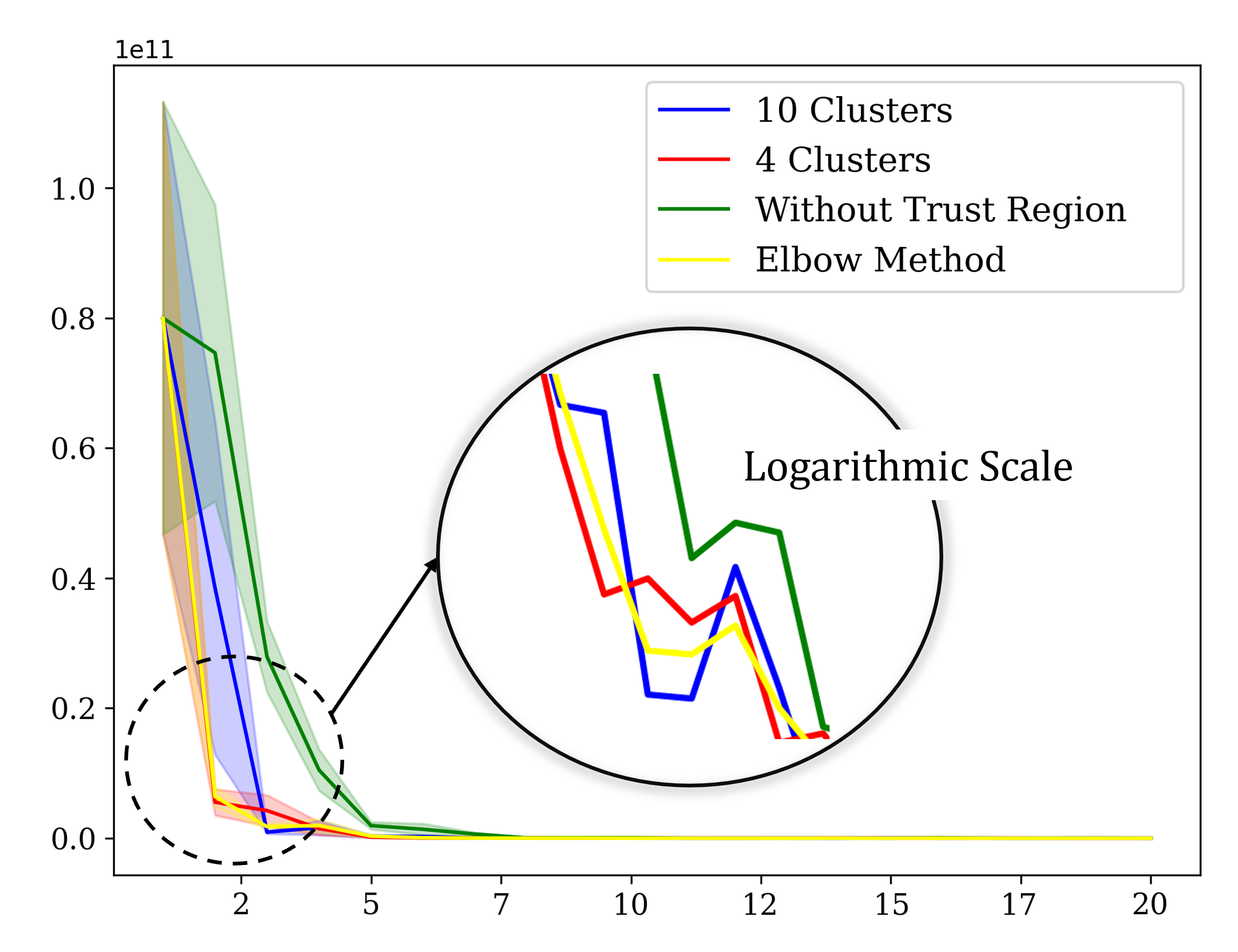}
    \put(-160,-5){\normalsize \rotatebox{0}{ Design Evaluations}}
    \put(-250,80){\normalsize\rotatebox{90}{EHVI}}
\caption{}
    \label{fig:Branin-log}
    \end{subfigure}
    \caption{The shaded region represents two standard errors of the mean across 20 replications for each test problem:(a) Bohachevsky-Sphere Function (b) Branin-Currin Function}
    \label{fig:Results with different Clustring}
\end{figure*}
Figure \ref{fig:Results with different Clustring} shows NOSTRA's performance with fixed and adaptive number of trust regions for the two test problems. Overall, NOSTRA outperforms methods without the use of trust regions, regardless of the chosen number of trust regions. Specifically, for the Bohachevsky-Sphere function (Figure \ref{fig:bohachovke}), using a fixed number of four clusters yields results comparable to those obtained with adaptive clustering via the Elbow Method.
For the Branin-Currin problem, NOSTRA also shows strong performance, as seen in Figure \ref{fig:Branin-log}. However, the behavior of the adaptive selection of the number of trust regions shows a discernible improvement on the Bohachevsky-Sphere function. Setting the number of clusters to ten results in a steeper initial decline in the EHVI graph. This occurs because smaller clusters allow NOSTRA to refine the trust regions more effectively, focusing the search on areas with the highest probability of containing the Pareto frontier restricting unnecessary exploration. In contrast, using a fixed four clusters or the Elbow Method allows for greater exploration of regions with higher probabilities of containing the Pareto frontier. These approaches prioritize sampling points that, while not necessarily optimal, have a reasonable chance of being included in the Pareto frontier. The differences between the fixed number of four trust regions and the adaptive Elbow Method in Figure \ref{fig:Branin-log} are attributed to the dynamic nature of the Elbow Method, where the number of clusters changes during the optimization process. Unlike the Bohachevsky-Sphere function, the Branin-Currin problem is less smooth, and the optima of its two objectives conflict significantly. Consequently, the number of clusters determined by the Elbow Method varies, highlighting the difference between this adaptive approach and the fixed four-clusters strategy.

\subsection{Evaluating Efficiency  of NOSTRA Across Varying Noise Levels in Initial Data}
\begin{figure}[htb]
    \centering
    \includegraphics[width=\linewidth]{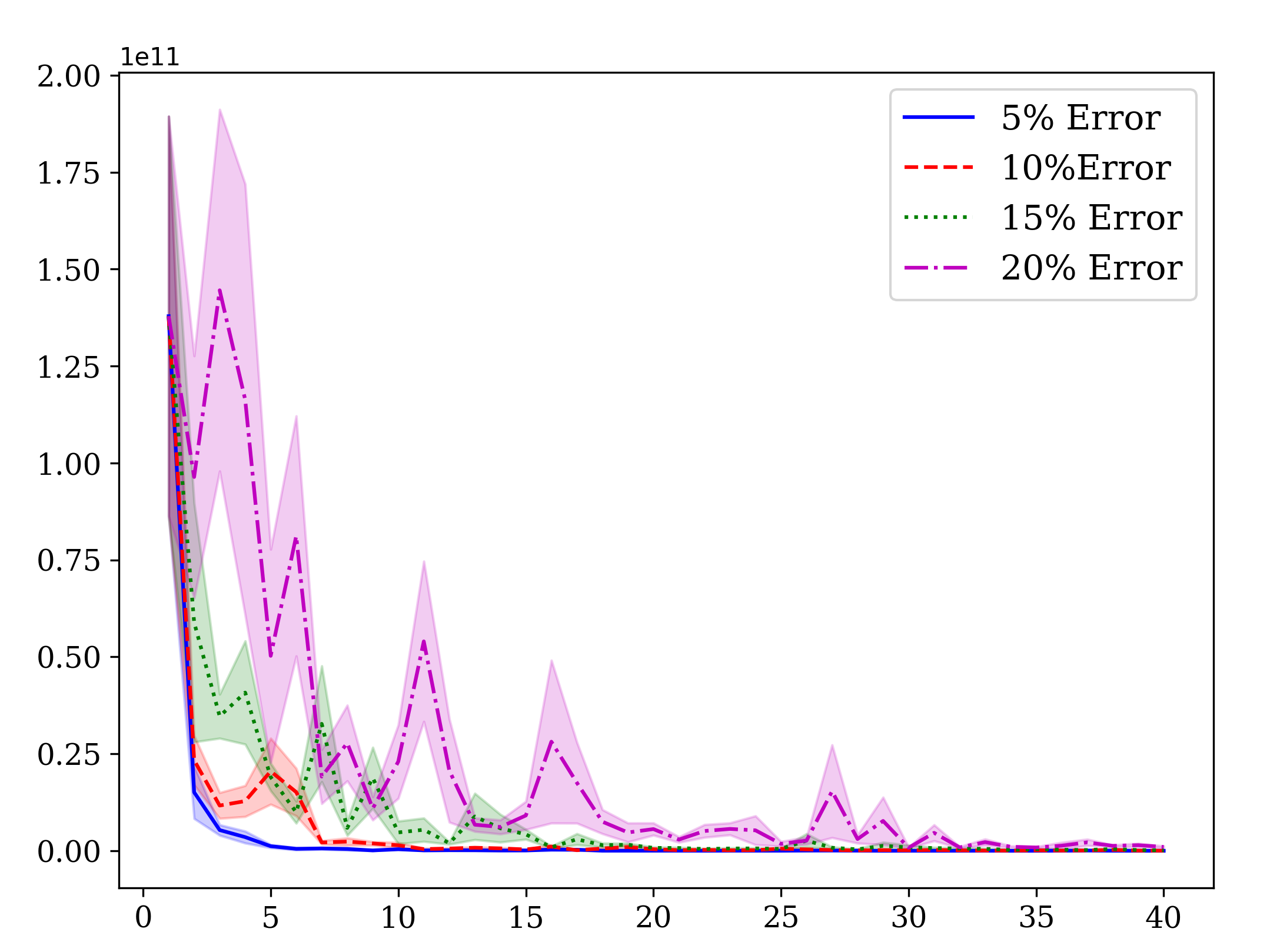}
    \put(-160,-10){\normalsize \rotatebox{0}{ Design Evaluations}}
    \put(-260,80){\normalsize\rotatebox{90}{EHVI}}
    \caption{Sequential Optimization Performance for Branin-Currin Function: Detailed Comparison in Logarithmic and Normal Scales}
    \label{fig:noiselevels}
\end{figure}

Here, we evaluated NOSTRA with clustering using the Elbow Method to assess its performance under varying noise levels. In this context, “noise level” refers to variability in the observed output values $(y)$, and is labeled as “Error” in Figure \ref{fig:noiselevels}. This noise is modeled as zero-mean Gaussian noise added to the observed data to simulate experimental uncertainty. The standard deviation of the noise is defined as a percentage of the overall output range of the function, with typical values set to 5\% ,10\%,15\% and 20\%.

As shown in Figure~\ref{fig:noiselevels}, the algorithm performs well when the noise level is low, with a standard deviation of  5\% relative to the range of the functions. Under this condition, NOSTRA readily identifies trust regions and directs the optimization toward the Pareto frontier. However, at higher levels of experimental uncertainty, such as 20\%, the algorithm begins to exhibit significant fluctuations, reflecting its difficulty in distinguishing between the mean of the function and the experimental uncertainty. This can result in large fluctuations in the posterior predictive distributions. While this requires additional experimental resources, we can observe a reliable trend in the reduction of the EHVI as more samples are added. Specifically, the adaptive framework of NOSTRA mitigates these issues through the use of trust regions that help guide the optimization process toward areas more likely to contain the Pareto frontier, even in the presence of experimental uncertainty. As illustrated in Figure \ref{fig:noiselevels}, the algorithm initially experiences large spikes during early iterations. This behavior arises from attempts to fit the surrogate model to noisy data, causing significant shifts in the location of the trust regions. This feature enhances the efficiency of the algorithm by mitigating the impact of high noise levels through the incorporation of prior knowledge GP construction and hyperparameter tuning. In high-noise regimes, GPs are prone to overfitting spurious fluctuations, which can result in numerical instability. By embedding prior knowledge into the GP kernel design, as described in Section \ref{Sec:priorInfo}, the algorithm reduces the influence of outliers and unreliable observations, thereby lowering the cost of optimization and improving posterior updates. However, as more data is gathered and the surrogate model stabilizes, the trust regions become more consistent, leading to smaller fluctuations and more reliable convergence. Ultimately, NOSTRA effectively balances exploring areas with large prediction uncertainties with focusing on trust regions, allowing it to maintain strong optimization performance.

\section{Concluding Remarks}\label{conclusion}

In this paper, we introduced NOSTRA, a method for efficient multi-objective Bayesian optimization in scenarios where data is sparse, scarce, and noisy due to experimental uncertainty. Such conditions are common in physical experiments (e.g., 3D-printed devices) and simulations (e.g., molecular dynamics and agent-based models). In addition, for these types of problems, data is often a costly commodity so only a small number of experiments can be conducted (e.g., less than ten samples per input variable). NOSTRA enhances MOBO by incorporating prior knowledge into the construction of Gaussian processes and by defining trust regions directly by considering the probability of a design being Pareto optimal. This allows the algorithm to focus exploration on areas with a higher probability of containing Pareto-optimal solutions, thereby accelerating convergence and reducing the number of required evaluations.

Our experiments demonstrated that increasing the number of trust regions can reduce the experimental budget but may risk excluding other potentially satisfactory solutions. This trade-off provides flexibility for users to balance convergence efficiency and solution diversity depending on specific design goals. While NOSTRA shows strong performance in low-dimensional spaces, its scalability to high-dimensional problems remains an open challenge. Future work will investigate strategies for improving scalability, including dimensionality reduction techniques and more efficient hyperparameter tuning. Additionally, We aim to extend NOSTRA to support batch sampling, model non-Gaussian and heteroscedastic noise structures, and integrate it with transfer learning systems for adaptive experimental design. Ultimately, we plan to validate the approach in practical applications where data is scarce, sparse, and uncertain, demonstrating NOSTRA’s practical relevance for robust and efficient optimization in complex design tasks.

\section{Acknowledgments}
This research was supported by Chang'an-Dublin International College of Transportation (CDIC).
\bibliographystyle{asmeconf}  
\bibliography{main}
\end{document}